%% file: neurips_2019.tex
\documentclass{article}

\usepackage[preprint,nonatbib]{neurips_2019}

\usepackage[utf8]{inputenc} % allow utf-8 input
\usepackage[T1]{fontenc}    % use 8-bit T1 fontshttps://www.overleaf.com/7341681675npqczwdqwtzv
\usepackage{hyperref}       % hyperlinks
\usepackage{url}            % simple URL typesetting
\usepackage{booktabs}       % professional-quality tables
\usepackage{amsfonts}       % blackboard math symbols
\usepackage{nicefrac}       % compact symbols for 1/2, etc.
\usepackage{microtype}      % microtypography

\usepackage{graphicx}
\usepackage{xcolor}
\usepackage{amsmath}
\usepackage{multirow}
\usepackage{graphbox}
\usepackage{makecell}

\usepackage{array}
\usepackage{arydshln}
\usepackage{float}

\usepackage[title]{appendix}

\title{KG-GAN: Knowledge-Guided Generative Adversarial Networks}

\author{
  Che-Han Chang$^1$, Chun-Hsien Yu$^1$, Szu-Ying Chen$^1$, and Edward Y. Chang$^{1,2}$ \\
  $^1$HTC Research \& Healthcare \\
  $^2$Stanford University \\
  \texttt{\{chehan\_chang,jimmy\_yu,claire.sy\_chen,edward\_chang\}@htc.com} \\
}
\graphicspath{{./imgs/}}

\begin{document}

\maketitle

\begin{abstract}

Can generative adversarial networks (GANs) generate roses of various colors given only
roses of red petals as input?  The answer is negative, since GANs' discriminator would reject
all roses of unseen petal colors.  In this study, we propose knowledge-guided GAN (KG-GAN) to
fuse domain knowledge with the GAN framework.  KG-GAN trains two
generators; one learns from data whereas the
other learns from knowledge with a constraint function.
Experimental results demonstrate
the effectiveness of KG-GAN in generating unseen flower categories
from seen categories given textual descriptions of the unseen ones.
\end{abstract}

\input{macros}
\input{defs}

\input{1_intro}

\input{2_related}
\input{3_method}
\input{4_experiments}

\input{5_conclusion}

\bibliographystyle{unsrt}
% \bibliography{refs}

\input{refs.bbl}
\begin{appendices}
\input{appendix_workshop/appendix.tex}
\end{appendices}

\end{document}

%% file: macros.tex
%% editing comment
\newcommand{\ignore}[1]{}   % ignore this
\newcommand{\cmt}[1]{\begin{sloppypar}\large\textcolor{red}{#1}\end{sloppypar}}
\newcommand{\note}[1]{\cmt{Note: #1}}

\newcommand{\todo}[1]{ \textcolor{red}{[{\bf TODO}: #1]}}
\newcommand{\torevise}[1]{\textcolor{blue}{#1}}
\newcommand{\copied}[1]{ \textcolor{red}{[COPIED: #1]}}
\newcommand{\frank}[1]{\textcolor{blue}{[Frank: #1]}}
\newcommand{\ed}[1]{\textcolor{blue}{[Ed: #1]}}

\newcommand{\comment}[1]{\textcolor{blue}{#1}}
\newcommand{\revised}[1]{\textcolor{blue}{#1}}

%% file: defs.tex
\newcommand{\ascalar}[1]{#1}
\newcommand{\avector}[1]{{\mathbf{#1}}}
\newcommand{\amatrix}[1]{\mathbf{#1}}
\newcommand{\aset}[1]{\mathbf{#1}}
\newcommand{\B}[1]{\mathbf{#1}}

\newcommand{\LL}{\mathcal{L}}
\newcommand{\E}{\mathbb{E}}
\newcommand{\Exp}[2]{\mathbb{E}_{#1} \left[ {#2} \right] }

\newcommand{\A}{a}
\newcommand{\sourceA}{\A}
\newcommand{\targetA}{\hat{\A}}
\newcommand{\RA}{v}
\newcommand{\coeff}{\alpha}

\newcommand{\adv}{\text{Real}}
\newcommand{\con}{\text{Match}}
\newcommand{\itp}{\text{Interp}}
\newcommand{\cyc}{\text{Cycle}}
\newcommand{\sel}{\text{Self}}
\newcommand{\cls}{\text{Cls}}

\newcommand{\Dadv}{D_\adv}
\newcommand{\Dcon}{D_\con}
\newcommand{\Dint}{D_\itp}

\newcommand{\Ladv}{\LL_\adv}
\newcommand{\Lcon}{\LL_\con}
\newcommand{\Lint}{\LL_\itp}
\newcommand{\Lcyc}{\LL_\cyc}
\newcommand{\Lsel}{\LL_\sel}
\newcommand{\Lcls}{\LL_\cls}

\newcommand{\w}{\lambda}

\newcommand{\zerovec}{\vec{0}}

\newcommand{\V}{\surd}

% target category, domain, label, class
\newcommand{\y}{y}

\newcommand{\kgganonehot}{\text{KG-GAN-o}}
\newcommand{\snganfeat}{\text{SN-GAN-v}}
\newcommand{\kgganfeat}{\text{KG-GAN-v}}

%% file: 1_intro.tex
% \vspace{-6pt}
\section{Introduction}
% \vspace{-6pt}

%%%%%%%%%%%%%%%%%%%%%%%%%%%%%%%%
% 1) GANs and cGANs
%%%%%%%%%%%%%%%%%%%%%%%%%%%%%%%%
Generative adversarial networks (GANs)~\cite{goodfellow2014generative} and their variants have received massive attention in the machine learning and computer vision communities recently due to their impressive performance in various tasks, such as categorical image generation~\cite{miyato2018cgans}, text-to-image synthesis~\cite{reed2016generative}~\cite{zhang2017stackgan}, image-to-image translation~\cite{isola2017image}~\cite{zhu2017unpaired}~\cite{choi2018stargan}, and semantic manipulation~\cite{park2019semantic}.
The goal of GANs or e.g., cGANs is to learn a generator that mimics the underlying distribution represented by 
a set of training data.
Considerable progress has been made to improve the robustness of GANs.

%%%%%%%%%%%%%%%%%%%%%%%%%%%%%%%%
% 2) training data is imperfect
%%%%%%%%%%%%%%%%%%%%%%%%%%%%%%%%

However, when the training data does not represent the {\em target concept} or underlying distribution,
i.e., the empirical training distribution deviates from the underlying distribution,
GANs trained from under-represented training data mimic the training distribution, but not the underlying one.

Training a GAN conditioned on category labels requires collecting training examples for each category.
If some categories are not available in the training data, then it appears infeasible to learn to generate their representations without any additional information.
Even when training a GAN model for one concept, 
if the training data cannot reflect all various
characteristics\footnote{One can treat each distinct characteristic, such as a rose color, as a category.} 
of that target concept, the generator cannot generate all diverse characteristics 
of that concept.
For instance, if the target concept is roses and training data consists of only red colored roses, 
the GANs' discriminators would 
reject the other colors of roses and fail to generate non-red colored roses.  At the same
time, we want to ensure that GANs will not generate a rose with an unnatural color.  Another example
is generating more stroke CT training images using GANs (given some labeled training instances)
to improve prediction accuracy.
If the training data covers only a subset of possible stroke characteristics (e.g.,
ischemic stroke (clots)), current GANs cannot help
generate unseen characteristics (e.g., hemorrhagic stroke (Bleeds)). 

In this paper,
we propose Knowledge-Guided Generative Adversarial Networks (KG-GANs), which incorporates \textit{domain knowledge} into GANs to enrich and expand the generated distribution, hence increasing the diversity of the generated data for a target concept.
Our key idea is to leverage domain knowledge as another learning source other than data 
to guide the generator to explore different regions of the image manifold.
By doing so, the generated distribution goes beyond the training distribution and better mimics the underlying one. 
A successful generator must consider knowledge to achieve
the twin goals of:
\begin{itemize}
\item Guiding the generator to explore diversity productively, and
\item Allowing the discriminator to tolerate diversity reasonably.
\end{itemize}
Note that domain knowledge serves GANs as a guide to not only explore diversity but also constrain 
exploring into regions that are knowingly impossible, such as generating gray roses.

%For instance, in the task of hair recoloring (or hair color transfer), 
%if we want to train an image-to-image translation model that recolors hair by rare colors such as purple, it is necessary to collect images with those hair colors.
%However, it is impractical to collect all possible dyed hair colors for arbitrary recoloring.

%This situation occurs because data collection is labor intensive and it is difficult to be thorough.
%Additionally, some modes of the underlying distribution could be missing in the training data 
%due to insufficient quantity and in particular, diversity.
%Specifically, we consider that the underlying distribution is under-represented by the training data \textit{at the category level}, i.e., some categories have no training data examples.

Our KG-GAN framework consists of two parts:
(1) constructing the domain-knowledge for the task at hand, and
(2) training two generators $G_1$ and $G_2$ that are conditioned on available and unavailable categories, respectively.
We formalize domain-knowledge as a constraint function that explicitly measures whether an image has the desired characteristics of a particular category.
On the one hand, the constraint function is task-specific and guides the learning of $G_2$.
On the other hand, we share weights between $G_1$ and $G_2$ to leverage the knowledge learned from available to unavailable categories.

%%%%%%%%%%%%%%%%%%%%%%%%%%%%%%%%
% 4) how to model knowledge into GANs.
%%%%%%%%%%%%%%%%%%%%%%%%%%%%%%%%

% \ignore{
% We validate KG-GAN on two tasks: shape generation and hair recoloring.
% \torevise{Figure 1 illustrates some examples of these two tasks.}
% For shape generation, given a training set consists of only odd-sided polygon images, we aim to train a side-conditioned generator that is capable of generating both odd- and even-sided polygon images.
% The domain knowledge we use here is that an n-sided polygon is characterized by the number of sides, which can be identified by the number of dominant gradient directions.
% The constraint function used here is a gradient-histogram-based neural network that performs polygon side estimation and can be trained from odd-sided polygon images while generalizing to even-sided ones.
% }

We validate KG-GAN on the task of generating flower images.
We aim to train a category-conditional generator that is capable of generating both 
seen and unseen categories.
The domain knowledge we use here is a semantic embedding representation that describes 
the semantic relationships among flower categories,
such as the textual features from descriptions of each category's appearance.
The constraint function used is a deep regression network that predicts the semantic 
embedding vector of the underlying category of an image.

%For hair recoloring, given the training data consists of three hair colors, our method trains an image-to-image translation model that is capable of recoloring face images with \textit{arbitrary} hair colors.
%We leverage the domain knowledge that hair color is characterized by the dominant color of the hair region
%(not including eyebrows and beard).
%In this case, hair segmentation plays a key role in implementing the constraint function that performs hair color estimation.
%We jointly train the segmentation network, $G_1$, and $G_2$ in an unsupervised manner.
%We additionally leverage the observation that hair recoloring can be safely assumed as a spatially invariant linear transformation applied only on the hair region.
%We propose a new generator architecture that outputs transformations rather than images.
%This significantly improves the segmentation accuracy and hence the recoloring quality.

Our main contributions are summarized as follows: 
\begin{enumerate}
\item We tackle the problem that the training data cannot well represent the underlying distribution.
\item We propose a novel generative adversarial framework that incorporates domain knowledge into GAN methods.
\item We demonstrate the effectiveness of our KG-GAN framework on unseen flower category generation.
\end{enumerate}
%We are able to enrich the diversity of the generated distribution by generating characteristics 
%not available in the original training data.

%% file: 2_related.tex
% \vspace{-6pt}
\section{Related work}
% \vspace{-6pt}

Since a comprehensive review of the related works on GANs is beyond the scope of the paper,
we only review representative works most related to ours.

\textbf{Generative Adversarial Networks.}
Generative adversarial networks (GANs)~\cite{goodfellow2014generative}
introduce an adversarial learning framework that jointly learns a discriminator and 
a generator to mimic a training distribution.
Conditional GANs (cGANs) extend GANs by conditioning on additional information such as category label~\cite{mirza2014conditional,miyato2018cgans}, text~\cite{reed2016generative}, 
or image~\cite{isola2017image, zhu2017unpaired, choi2018stargan}.
SN-GAN~\cite{miyato2018cgans, miyato2018spectral} proposes a projection discriminator and a spectral normalization method to improve the robustness of training.
CycleGAN~\cite{zhu2017unpaired} employs a cycle-consistency loss to regularize the generator for unpaired image-to-image translation.
StarGAN~\cite{choi2018stargan} proposes an attribute-classification-based method that adopts a single generator for multi-domain image-to-image translation.
Hu et al.~\cite{hu2018deep} introduce a general framework that incorporates domain knowledge into deep generative models.
Its primary purpose is improving quality, but not improving diversity, which is our goal.

\textbf{Diversity.}
The creative adversarial network (CAN)~\cite{elgammal2017can} augments GAN with two style-based losses to make its generator go beyond the training distribution and thus generate diversified artistic images.
The imaginative adversarial network (IAN)~\cite{hamdi2019ian} proposes a two-stage generator that goes beyond a source domain (human face) and towards a target domain (animal face).
Both works aim to go beyond the training data.
However, they target artistic image generation and do not possess a well-defined underlying distribution.
Mode-Seeking GAN~\cite{mao2019mode} proposes a mode seeking regularization method to alleviate the mode collapse problem in cGANs, which happens when the generated distribution cannot well represent the training distribution.
Our problem appears similar but is ultimately different.
We tackle the problem in which the training distribution under-represents the underlying distribution.

\textbf{Zero-shot Learning}
We refer readers to~\cite{xian2017zero} for a comprehensive introduction and evaluation of representative zero-shot learning methods.
The crucial difference between the zero-shot method 
and our method is that they focus on image classification while we focus on image generation.
Recently, some zero-shot methods~\cite{xian2018feature, felix2018multi, xian2019f} proposed 
learning feature generation of unseen categories for training zero-shot classifiers.
Instead, this work aims to learn image generation of unseen categories.

%Specifically, increasing diversity at the category level can also be seen as conditional image generation under the zero-shot setting.
% f-CLSWGAN~\cite{xian2018feature}, Cycle-CLSWGAN~\cite{felix2018multi}, f-VAEGAN-D2~\cite{xian2019f}.

%% file: 3_method.tex
% \vspace{-6pt}
\section{KG-GAN}
% \vspace{-6pt}

\begin{figure*}
\begin{center}
	\includegraphics[width=1.00\linewidth]{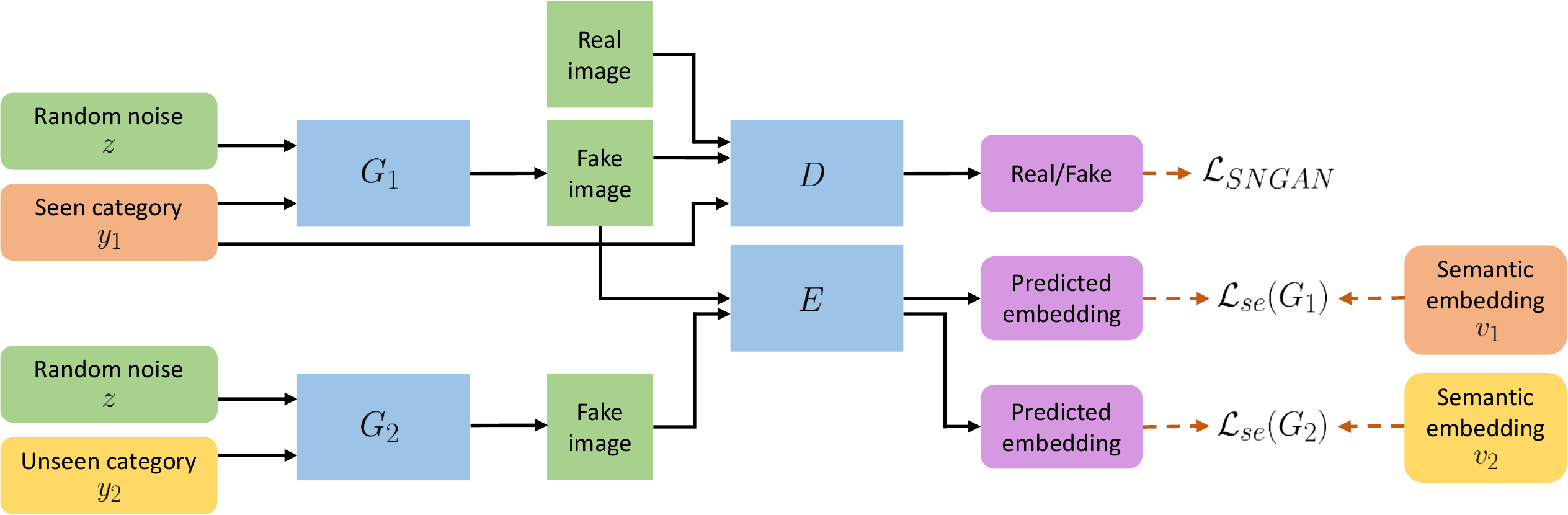}
% 	\vspace{-14pt}
% 	\setlength{\belowcaptionskip}{-0.6cm}
	\caption{
	The schematic diagram of KG-GAN for unseen flower category generation.
	There are two generators $G_1$ and $G_2$, a discriminator $D$, and an embedding regression network $E$ as the constraint function $f$.
	We share all the weights between $G_1$ and $G_2$.
	By doing so, our method here can be treated as training a single generator with a category-dependent loss that seen and unseen categories correspond to optimizing two losses ($\LL_{SNGAN}$ and $\LL_{se}$) and a single loss ($\LL_{se}$), respectively, where $\LL_{se}$ is the semantic embedding loss.
	}
	\label{fig:flower_overview}
\end{center}
\end{figure*}

This section presents our proposed KG-GAN that incorporates domain knowledge into the GAN framework.
%We first provide an overview of KG-GAN.
%Then, we show how to apply KG-GAN on two representative tasks:
%fine-grained image generation and hair recoloring.
We consider a set of training data under-represented at the category level, i.e.,
all training samples belong to the set of seen categories, denoted as $Y_1$ (e.g.,
red category of roses),
while another set of unseen categories, denoted as $Y_2$ (e.g., any other
color categories), has no training samples.
Our goal is to learn categorical image generation for both $Y_1$ and $Y_2$.
To generate new data in $Y_1$, KG-GAN applies an existing GAN-based method to train a category-conditioned generator $G_1$ by minimizing GAN loss $\LL_{GAN}$ over $G_1$.
To generate unseen categories $Y_2$, KG-GAN trains another generator $G_2$ from the domain knowledge,
which is expressed by a constraint function $f$ that explicitly measures whether an image has the desired characteristics of a particular category.

KG-GAN consists of two parts:
(1) constructing the domain knowledge for the task at hand, and
(2) training two generators $G_1$ and $G_2$ that condition on available and unavailable categories, respectively.
KG-GAN shares the parameters between $G_1$ and $G_2$ to couple them together and to transfer knowledge learned from $G_1$ to $G_2$.
Based on the constraint function $f$, KG-GAN adds a \textit{knowledge loss}, denoted as $\LL_{K}$, to train $G_2$.
The general objective function of KG-GAN is written as
$\min_{G_1, G_2} \LL_{GAN}(G_1) + \lambda\LL_{K}(G_2)$.

%%%%%%%%%%%%%%%%%%%%%%%%%%%%%%%%%%%%%%%%%%%%%%%%%%%%%%%%%%%%%%%%%%%%%%%%%%%%%%%%%%%%%%%%%

%\vspace{-8pt}
%\subsection{Fine-grained Image Generation}
%\label{sec:3_2}
%\vspace{-6pt}

Given a flower dataset in which some categories are unseen,
our aim is using KG-GAN to generate unseen categories in addition to the seen categories.
Figure~\ref{fig:flower_overview} shows an overview of KG-GAN for unseen flower category generation.
Our generators take a random noise $z$ and a category variable $\y$ as inputs and generate an output image $x'$.
In particular,
    $G_1: (z,\y_1) \mapsto x_1'$
and $G_2: (z,\y_2) \mapsto x_2'$,
where $\y_1$ and $\y_2$ belong to the set of seen and unseen categories, respectively.

We leverage the domain knowledge that each category is characterized by a semantic embedding representation,
which describes the semantic relationships among categories.
In other words, we assume that each category is associated with a semantic embedding vector $v$.
For example, we can acquire such feature representation from the textual descriptions of each category.
(Figure~\ref{fig:dataset} shows example textual descriptions for four flowers.)
We propose the use of the semantic embedding in two places.
One is for modifying the GAN architecture,
and the other is for defining the constraint function.
(Using the Oxford flowers dataset, we show how semantic embedding is done
in Section~\ref{sec:exp}.)

%We characterize fine-grained categories by a semantic embedding representation.
% Specifically, the category variable $\y$ can be defined in two ways:
% (1) a one-hot vector indicating which target category, or
% (2) the semantic embedding vector of the target category.
% We experiment with both choices and show the results in Section~\ref{sec:exp}.

% SN-GAN loss
%\textbf{SN-GAN.}
KG-GAN is developed upon SN-GAN~\cite{miyato2018cgans, miyato2018spectral}.
SN-GAN uses a projection-based discriminator $D$ and adopts spectral normalization for discriminator regularization.
The objective functions for training $G_1$ and $D$ use a hinge version of adversarial loss.
The category variable $\y_1$ in SN-GAN is a one-hot vector indicating which target category.
KG-GAN replaces the one-hot vector by the semantic embedding vector $v_1$.
By doing so, we directly encode the similarity relationships between categories into the GAN training.

The loss functions of the modified SN-GAN are defined as
\begin{equation}
    \begin{split}
        & \LL_{SNGAN}^G(G_1) = -\mathbb{E}_{z,v_1}[D(G_1(z,v_1),v_1)],~\text{and} \\
        & \LL_{SNGAN}^D(D) = \mathbb{E}_{x, v_1}[\max(0,1-D(x,v_1))]
                         + \mathbb{E}_{z, v_1}[\max(0,1+D(G_1(z,v_1),v_1))].
    \end{split}
\end{equation}
% \vspace{-8pt}

% domain knowledge
\textbf{Semantic Embedding Loss.}
We define the constraint function $f$ as predicting the semantic embedding vector of the underlying category of an image.
To achieve that, we implement $f$ by training an embedding regression network $E$ from the training data.
Once trained, we fix its parameters and add it to the training of $G_1$ and $G_2$.
In particular, we propose a semantic embedding loss $\LL_{se}$ as the role of knowledge loss in KG-GAN.
This loss requires the predicted embedding of fake images to be close to the semantic embedding of target categories.
$\LL_{se}$ is written as
\begin{equation}
    \begin{split}
        \LL_{se}(G_i) & = \mathbb{E}_{z, v_i} || E(G_i(z, v_i)) - v_i ||^2, \text{where}~i \in \{1, 2\}.
    \end{split}
% \vspace{-8pt}
\end{equation}

% total loss
\textbf{Total Loss.}
The total loss is a weighted combination of $\LL_{SNGAN}$ and $\LL_{se}$.
The loss functions for training $D$ and for training $G_1$ and $G_2$ are respectively defined as
\begin{equation}
    \begin{split}
        \LL^D & = \LL_{SNGAN}^D(D),~\text{and} \\
        \LL^G & = \LL_{SNGAN}^G(G_1) + \w_{se} (\LL_{se}(G_1) + \LL_{se}(G_2)).
    \end{split}
\end{equation}

%% file: 4_experiments.tex
\section{Experiments} \label{sec:exp}
% \vspace{-6pt}
%%%%%%%%%%%%%%%%%%%%%%%%%%%%%%%%%%%%%%%%%%%%%%%%%%%%%%%%%%%%
% Fine-grained experiments
%%%%%%%%%%%%%%%%%%%%%%%%%%%%%%%%%%%%%%%%%%%%%%%%%%%%%%%%%%%%
%\subsection{Fine-grained Image Generation.}
%\vspace{-6pt}

\begin{figure*}
\begin{center}
	\includegraphics[width=1.00\linewidth]{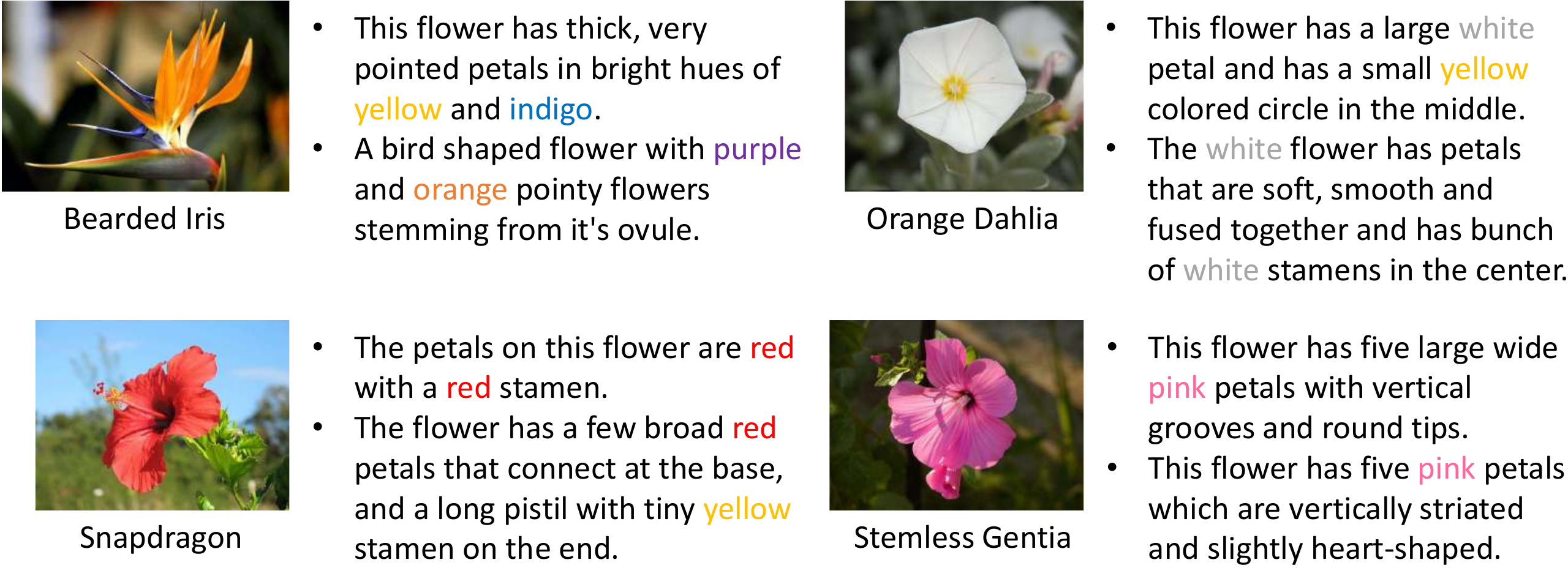}
% 	\vspace{-14pt}
% 	\setlength{\belowcaptionskip}{-0.6cm}
	\caption{
	\textbf{Oxford flowers dataset}. Example images and their textual descriptions.
	}
	\label{fig:dataset}
\end{center}
\end{figure*}

\begin{figure*}
\begin{center}
	\includegraphics[width=1.00\linewidth]{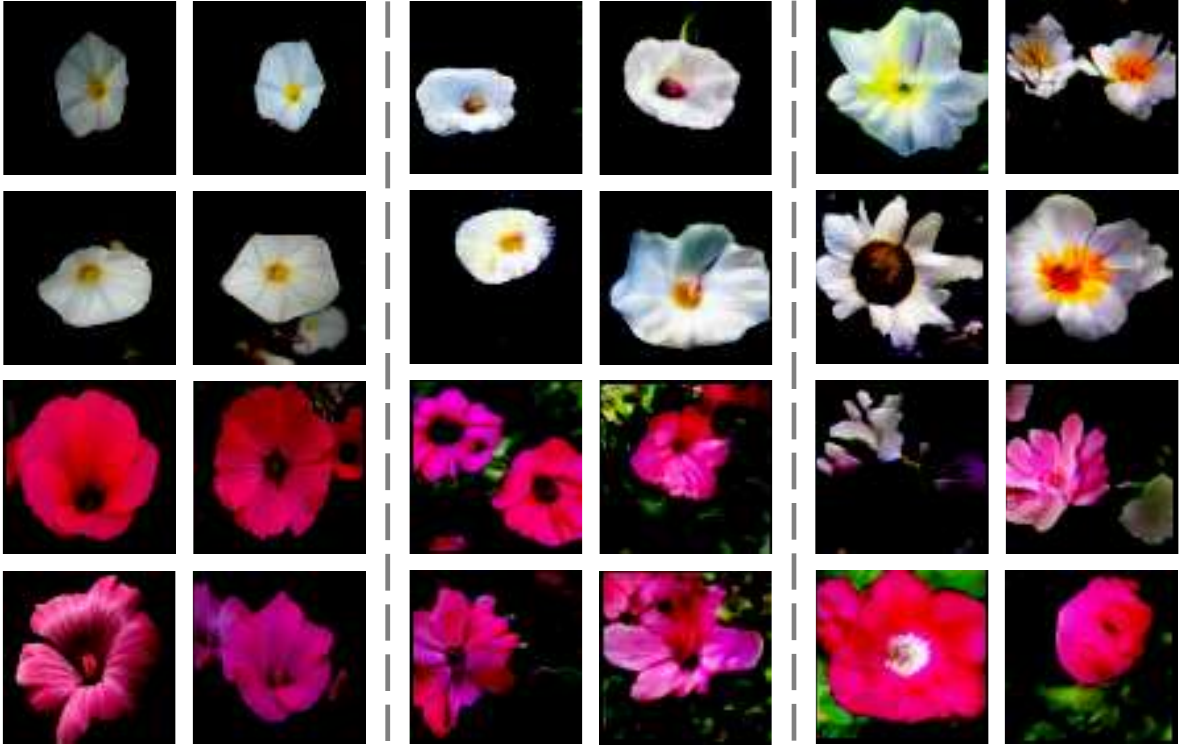}
% 	\vspace{-14pt}
% 	\setlength{\belowcaptionskip}{-0.6cm}
    \caption{\textbf{Unseen flower category generation.} Qualitative comparison between real images and the generated images from KG-GAN.
    \textit{Left}: Real images.
    \textit{Middle}: Successful examples of KG-GAN.
    \textit{Right}: Unsuccessful examples of KG-GAN.
    The top two and the bottom two rows are Orange Dahlia and Stemless Gentian, respectively.
    }
	\label{fig:flower_results}
\end{center}
\end{figure*}

\textbf{Experimental Settings.}
We use the Oxford flowers dataset~\cite{nilsback2008automated},
which contains $8{,}189$ flower images from $102$ categories (e.g., bluebell, daffodil, iris, and tulip).
Each image is annotated with $10$ textual descriptions.
Figure~\ref{fig:dataset} shows two representative descriptions for four flowers.
Following~\cite{reed2016learning}, we randomly split the images into $82$ seen and $20$ unseen categories.
To extract the semantic embedding vector of each category,
we first extract sentence features from each textual description using the fastText library~\cite{bojanowski2017enriching},
which takes a sentence as input and outputs a $300$-dimensional real-valued vector in range $[0, 1]$.
% \footnote{Garden Phlox description:
% Flowers	may	be pale blue, violet, pink, bright red, or white. Many are fragrant.
% Fertilized flowers typically
% produce	one	relatively large seed.}
Then we average over the features within each category to obtain the per-category feature
vector as the semantic embedding.
We resize the images to $64\times 64$ as the image size in our experiments.
For the SN-GAN part of the model, we use its default hyper-parameters and training configurations.
In particular, we train for $200$k iterations.
For the knowledge part, we use $\w_{se}=0.1$ in our experiments.

\textbf{Comparing Methods.}
We compare with SN-GAN trained on the full Oxford flowers dataset,
which potentially represents a performance upper-bound of our method.
Besides, we additionally evaluate two ablations of KG-GAN:
(1) One-hot KG-GAN: $y$ is a one-hot vector that represents the target category.
(2) KG-GAN w/o $\LL_{se}$: our method without $\LL_{se}$.

% We evaluate the following two versions of our method as mentioned in Section~\ref{sec:3_2}:
% (1) \kgganonehot: $y$ is a one-hot vector that represents the target category.
% (2) \kgganfeat: $y$ is the semantic embedding vector $v$ of the target category.
% For evaluation, we compare our generated images with the ones from the following:
% (1) the Oxford Flowers dataset,
% (2) SN-GAN trained on the full Oxford Flowers dataset,
% which represents a reasonable performance upper-bound of our method.
% (3) \snganfeat~trained on the full Oxford Flowers dataset, which is a variant of SN-GAN that, similar to \kgganfeat, takes $v$ as the target category.

\begin{figure}[htbp]
\tabcolsep=1pt
\renewcommand\arraystretch{5.1}
\begin{center}
    \begin{tabular}{c c c c c}
        \scriptsize{Real images} & \scriptsize{SN-GAN} & \scriptsize{One-hot KG-GAN} & \scriptsize{KG-GAN w/o $\LL_{se}$} & \scriptsize{KG-GAN} \\  [-16pt]
        \includegraphics[align=c, width=0.2\linewidth]{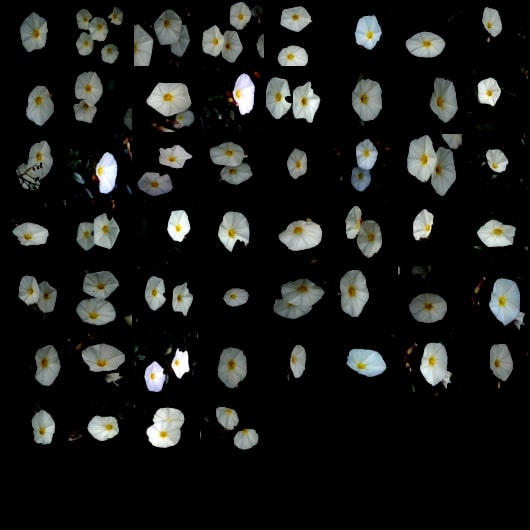}
        & \includegraphics[align=c, width=0.2\linewidth]{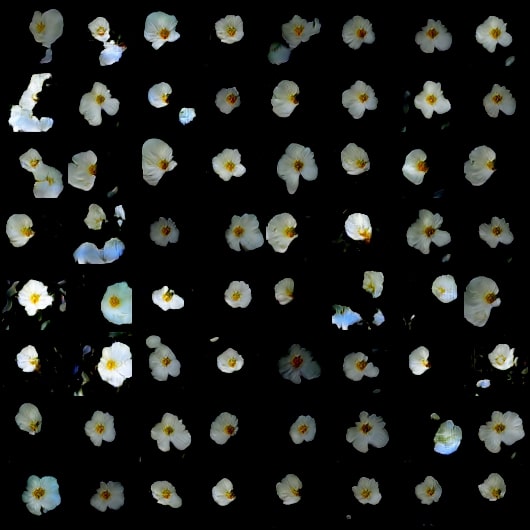}
        & \includegraphics[align=c, width=0.2\linewidth]{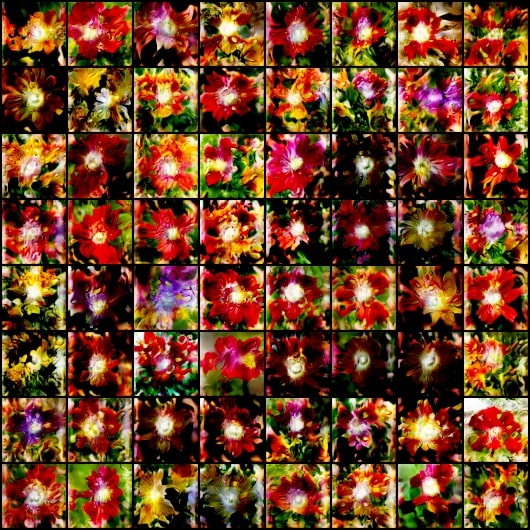}
        & \includegraphics[align=c, width=0.2\linewidth]{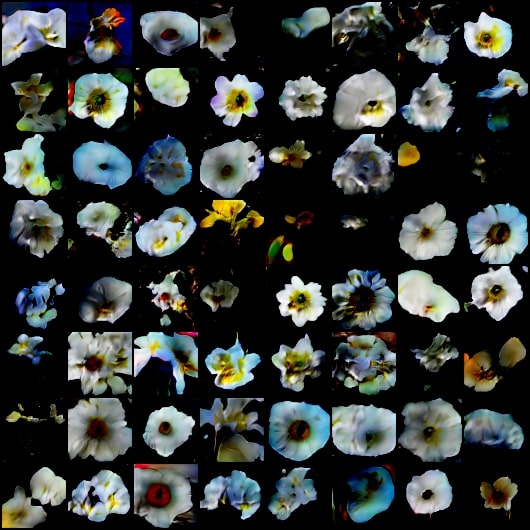}
        & \includegraphics[align=c, width=0.2\linewidth]{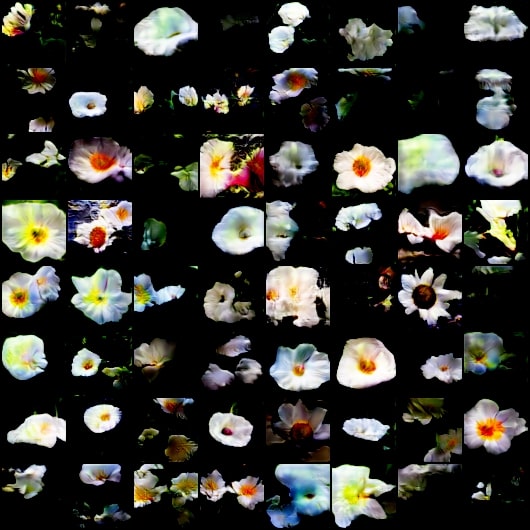} \\

        \includegraphics[align=c, width=0.2\linewidth]{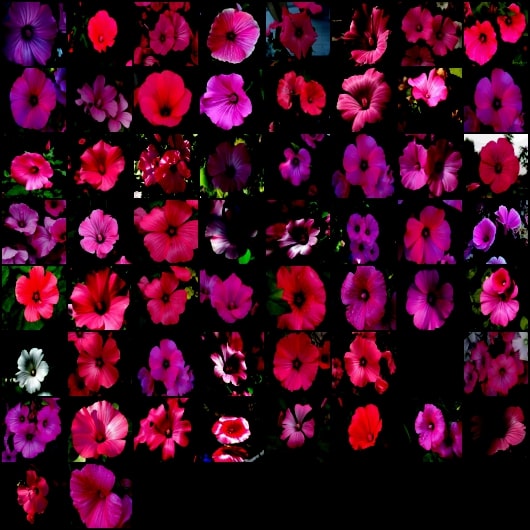}
        & \includegraphics[align=c, width=0.2\linewidth]{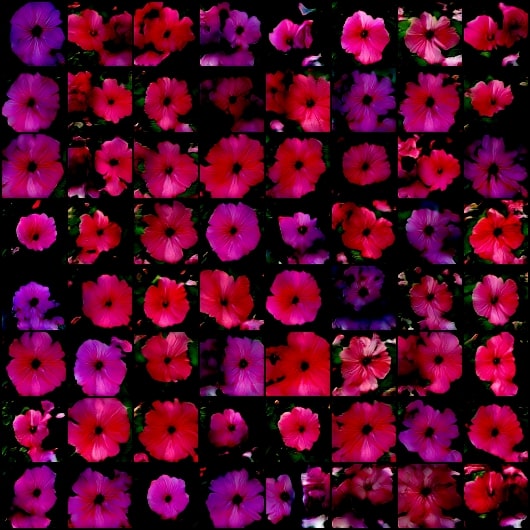}
        & \includegraphics[align=c, width=0.2\linewidth]{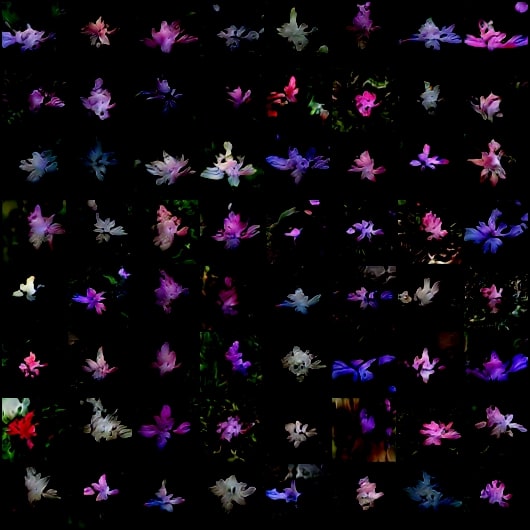}
        & \includegraphics[align=c, width=0.2\linewidth]{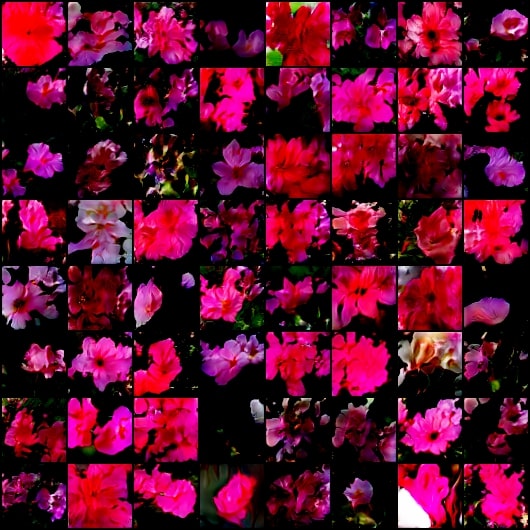}
        & \includegraphics[align=c, width=0.2\linewidth]{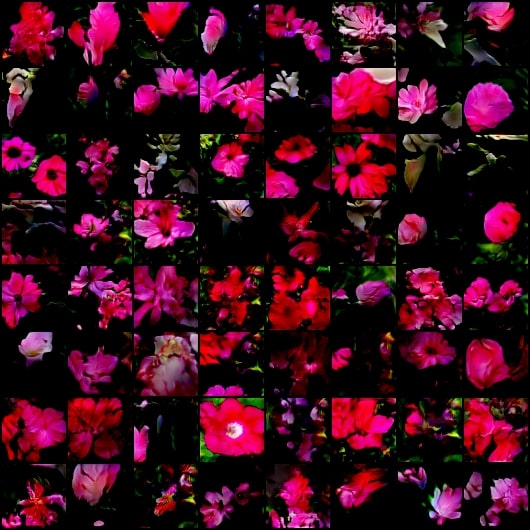} \\

        \includegraphics[align=c, width=0.2\linewidth]{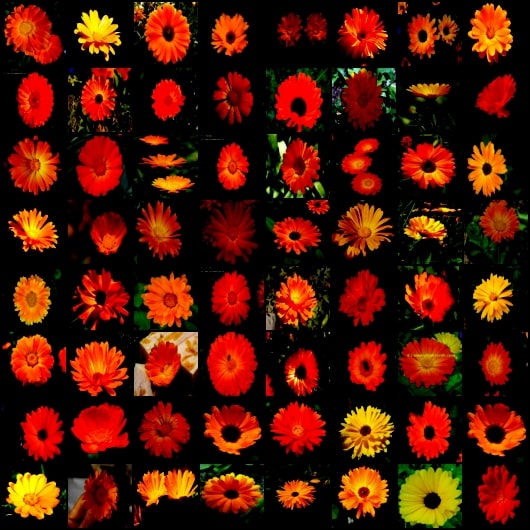}
        & \includegraphics[align=c, width=0.2\linewidth]{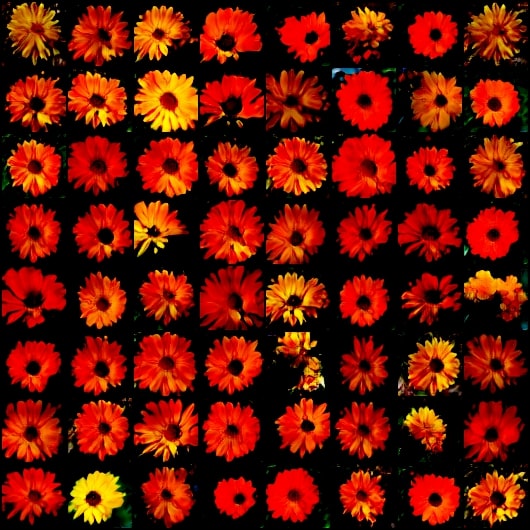}
        & \includegraphics[align=c, width=0.2\linewidth]{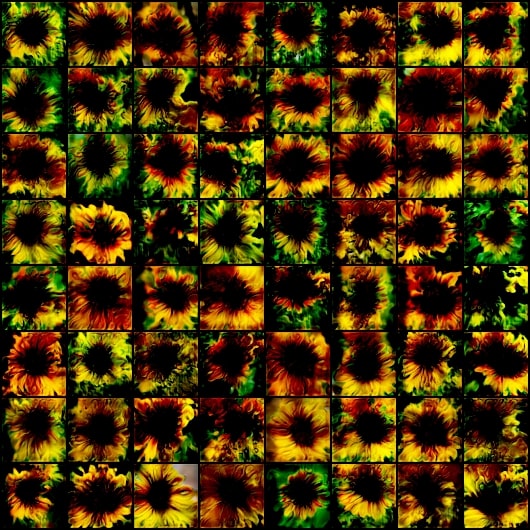}
        & \includegraphics[align=c, width=0.2\linewidth]{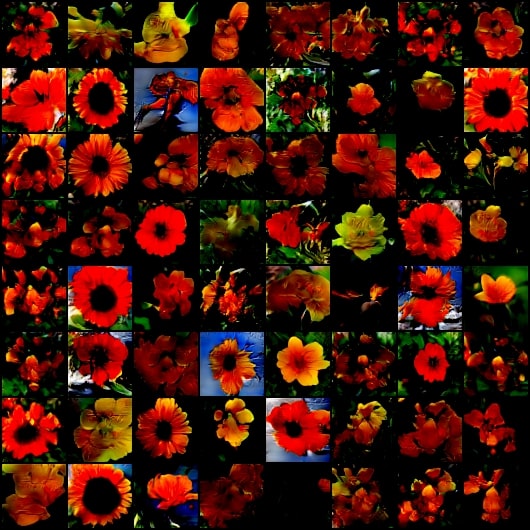}
        & \includegraphics[align=c, width=0.2\linewidth]{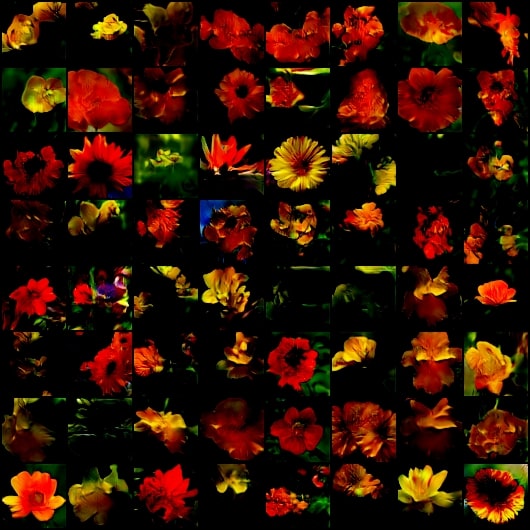} \\

        \includegraphics[align=c, width=0.2\linewidth]{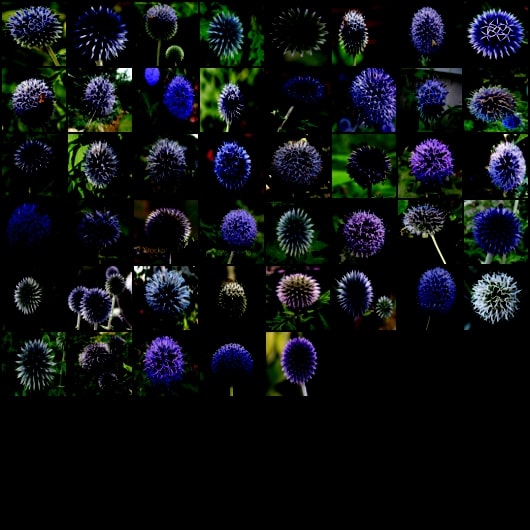}
        & \includegraphics[align=c, width=0.2\linewidth]{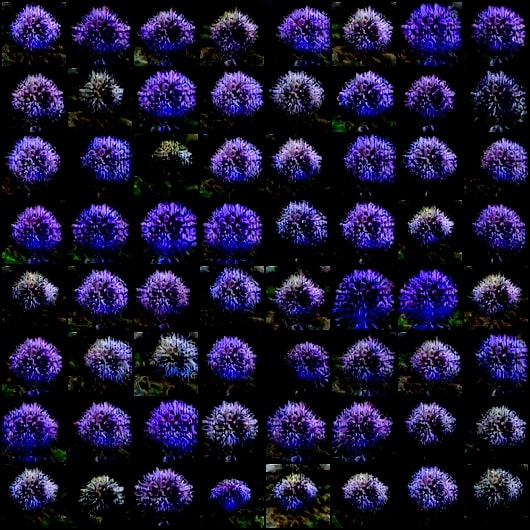}
        & \includegraphics[align=c, width=0.2\linewidth]{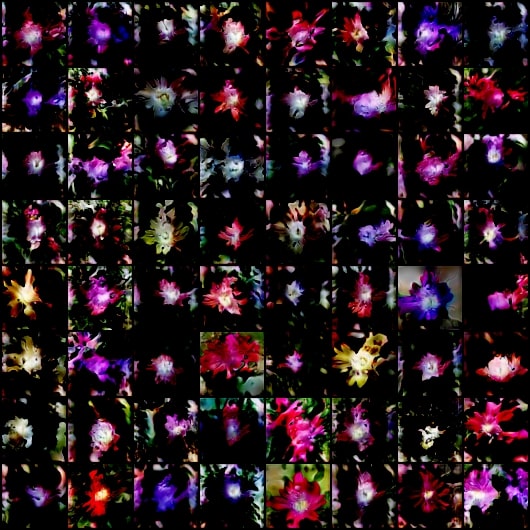}
        & \includegraphics[align=c, width=0.2\linewidth]{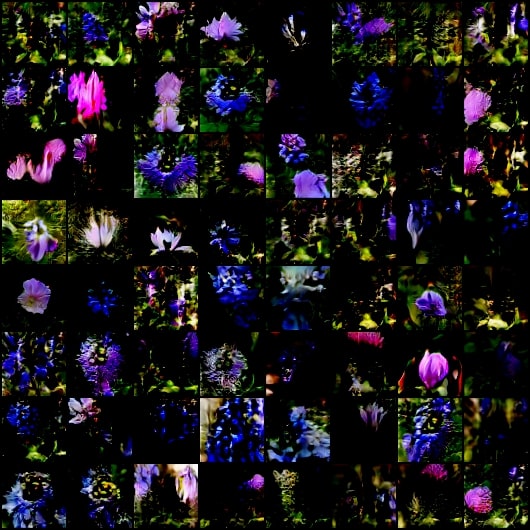}
        & \includegraphics[align=c, width=0.2\linewidth]{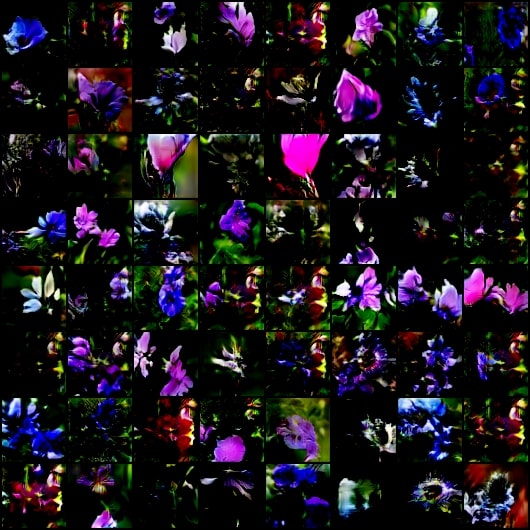} \\

        \includegraphics[align=c, width=0.2\linewidth]{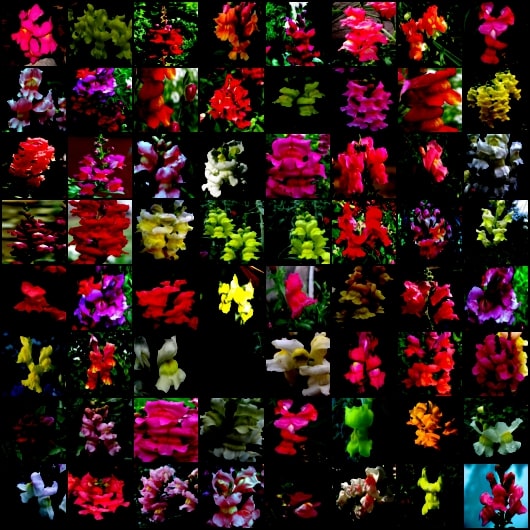}
        & \includegraphics[align=c, width=0.2\linewidth]{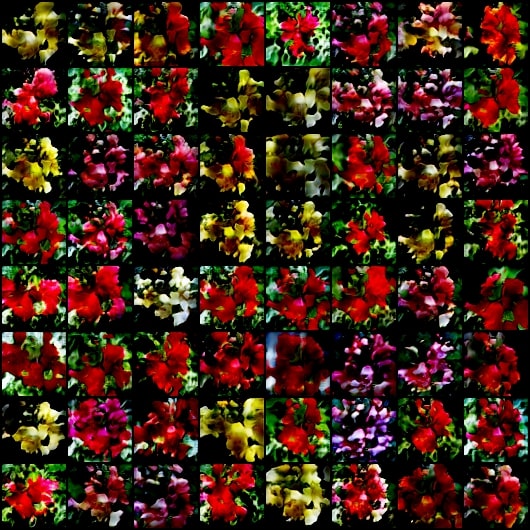}
        & \includegraphics[align=c, width=0.2\linewidth]{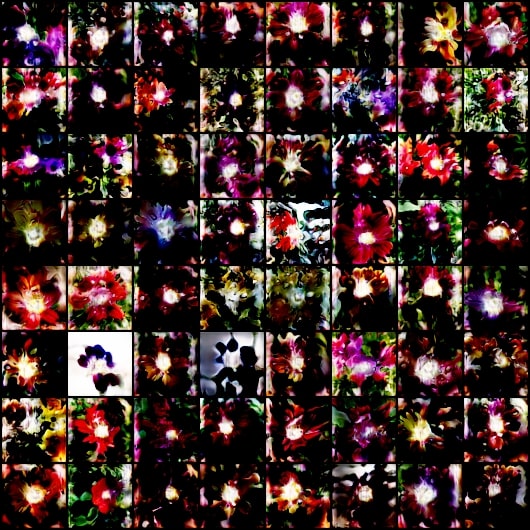}
        & \includegraphics[align=c, width=0.2\linewidth]{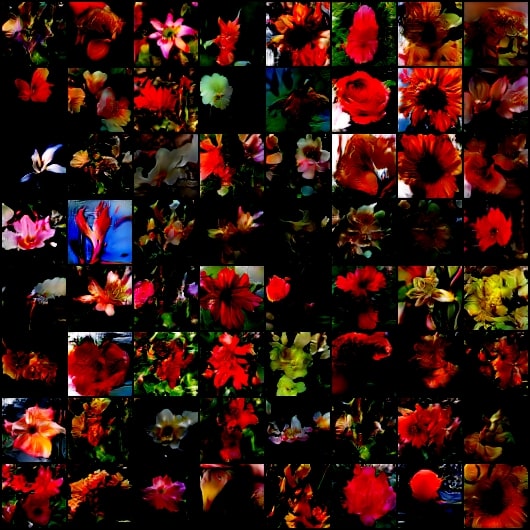}
        & \includegraphics[align=c, width=0.2\linewidth]{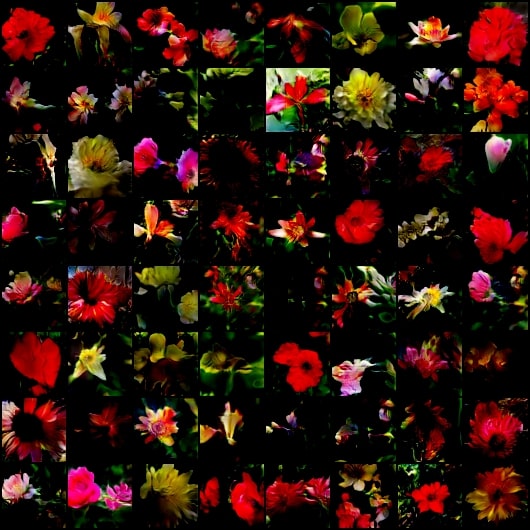} \\
    \end{tabular}
    \caption{Orange Dahlia, Stemless Gentian, Ball Moss, Bird of Paradise, and Bishop of Llandaff. Please zoom in to see the details.}
    \label{fig:flower0}
\end{center}
\end{figure}

\textbf{Results.} To evaluate the quality of the generated images,
we compute the FID scores~\cite{heusel2017gans} in a per-category manner as in~\cite{miyato2018cgans}.
Then, we average over the FID scores of the set of the seen and the unseen categories, respectively.
Table~\ref{tab:part1_fid} shows the seen and the unseen FID scores.
We can see from the table that in terms of the category condition,
semantic embedding gives better FID scores than one-hot representation.
Our full method achieves the best FID scores.
In Figure~\ref{fig:flower_results},
we show example results of two representative unseen categories.
In Figure~\ref{fig:flower0},
we show a visual comparison between real images, SN-GAN, and KG-GANs on $5$ unseen categories.
For the results of the remaining $15$ unseen categories, please refer to the Appendix.

% \textcolor{red}{Two enhancements are needed.  One, can you explain the ablation test in greater
% details such as what is and why one-hot.  Two, Figure 3 is still too small to see details.  Can you
% print just four well-generated photos and four poor ones, two for each category?
% Zooming into Figure 3 is useless.  Please
% generate high-res pictures like the ones in Figure 2.}

\textbf{Observations.}
From Table~\ref{tab:part1_fid} we have two observations.
First,
KG-GAN (conditioned on semantic embedding) performs better than One-hot KG-GAN.
This is because
One-hot KG-GAN learns domain knowledge only from the knowledge constraint while KG-GAN additionally learns the similarity relationships between categories through the semantic embedding as the condition variable.
Second,
when KG-GAN conditions on semantic embedding,
KG-GAN without $\LL_{se}$ still works.
This is because KG-GAN learns how to \textit{interpolate} among seen categories to generate unseen categories.
For example,
if an unseen category is close to a seen category in the semantic embedding space,
then their images will be similar.

As we can see from Figure~\ref{fig:flower_results},
our model faithfully generates flowers with the right color, but does not perform as well
in shapes and structures.
The reasons are twofold.
First, colors can be more consistently articulated on a flower.  Even if
some descriptors annotate a flower as red and while others annotate it as pink,
we can obtain a relatively consistent color depiction over, say,
ten descriptions.  Shapes and structures do not enjoy
as confined a vocabulary set as colors do.  In addition, the flowers in the same category may
have various shapes and structures due to aging and camera angles.
Since each image has $10$ textual descriptions and each category has an average number of $80$ images,
the semantic embedding vector of each category is obtained from taking an average
over about $800$ fastText feature vectors.
This averaging operation preserves the color information quite well while blurring the other aspects.
A better semantic embedding representation that encodes richer textual
information about a flower category is left as future work.

In Figures~\ref{fig:flower0},~\ref{fig:flower1},~\ref{fig:flower2},
and~\ref{fig:flower3} (Figures~\ref{fig:flower1} to \ref{fig:flower3} are
presented in the appendix), we show visual comparisons of SN-GAN and our method on
the Oxford flowers dataset.
We show both the real and the generated images of the following $20$ categories that
are seen to SN-GAN while unseen to KG-GAN:
Ball Moss,
Bird of Paradise,
Bishop of Llandaff,
Bougainvillea,
Canterbury Bells,
Cape Flower,
Common Dandelion,
Corn Poppy,
Daffodil,
Gaura,
GlobeThistle
Great Masterwort,
Marigold,
Mexican Aster,
Mexican Petunia,
Orange Dahlia,
Ruby-lipped Cattleya,
Stemless Gentian,
Thorn Apple, and
Trumpet Creeper.

\begin{table}
    \small
    \caption{Per-category FID scores of SN-GAN and KG-GANs.}
    \label{tab:part1_fid}
    \centering
    \begin{tabular}{lcccccc}
    \toprule
    Method & Training data & Condition & $\LL_{se}$ & Seen FID & Unseen FID \\
    \midrule
    SN-GAN                & $Y_1\cup Y_2$ & One-hot   &            & 0.6922   & 0.6201 \\ % ablation1
    \midrule
    One-hot KG-GAN        & $Y_1$         & One-hot   & $\V$       & 0.7077   & 0.6286 \\ % ablation2
    KG-GAN w/o $\LL_{se}$ & $Y_1$         & Embedding &            & 0.1412   & 0.1408 \\ % ablation4
    KG-GAN                & $Y_1$         & Embedding & $\V$       & $\mathbf{0.1385}$ & $\mathbf{0.1386}$ \\ % ablation5
    \bottomrule
    \end{tabular}
    %\kgganfeat & Embedding & $Y_1\cup Y_2$    &            & 0.1406    & 0.1369 \\ % ablation3
\end{table}

%% file: 5_conclusion.tex
% \vspace{-12pt}
\section{Conclusion}
% \vspace{-8pt}

We presented KG-GAN, the first framework that incorporates domain-knowledge 
into the GAN framework for improving diversity.
We applied KG-GAN on generating unseen flower categories.
Our results show that 
when the semantic embedding only provides coarse knowledge about a particular aspect of flowers,
the generation of the other aspects mainly borrows from seen classes,
meaning that there is still much room for improvement in knowledge representation.
%For hair recoloring, unsupervised segmentation plays an essential role in knowledge.
%Transformation generator improves segmentation accuracy while cycle-consistency further improves realism.
%KG-GAN takes the first step towards increasing diversity with knowledge.
%We hope that our work could inspire future research along the direction of knowledge-guided image generation.

%%%%%%%%%%%%%%%%%%%%%%%%%%%%%%%%%%%%%%%%%%%%%%%%%%%%%%%%%%%%%%%%%%%%%%%%%%%%%%%%%%%%%%%%%%
\ignore{
% stargan's discriminator aims at evaluating realism, which consider class info naturally.
% if we can disentangle the content into class-invariant one and class-specific one,
% then we can define a discriminator that only focus on class-invariant content 
% and that will be beneficial.
}

%% file: appendix_workshop/appendix.tex
\newpage
\section{More Results}

% \begin{figure}
% %\vspace{-14pt}
% \setlength{\belowcaptionskip}{-0.6cm}
% \tabcolsep=2pt
% \renewcommand\arraystretch{5}
% \begin{center}
%     \begin{tabular}{c c c c c c}
%         \scriptsize{Real images} & \scriptsize{SN-GAN} & \scriptsize{One-hot KG-GAN} & \scriptsize{KG-GAN w/o $\LL_{se}$} & \scriptsize{KG-GAN} \\  [-15pt]
%         \includegraphics[align=c, width=0.16\linewidth]{imgs/flower/unseen_64.jpg}
%         & \includegraphics[align=c, width=0.16\linewidth]{imgs/flower/1_unseen_64.jpg}
%         & \includegraphics[align=c, width=0.16\linewidth]{imgs/flower/2_unseen_64.jpg}
%         & \includegraphics[align=c, width=0.16\linewidth]{imgs/flower/4_unseen_64.jpg}
%         & \includegraphics[align=c, width=0.16\linewidth]{imgs/flower/5_unseen_64.jpg} \\

%         \includegraphics[align=c, width=0.16\linewidth]{imgs/flower/unseen_86.jpg}
%         & \includegraphics[align=c, width=0.16\linewidth]{imgs/flower/1_unseen_86.jpg}
%         & \includegraphics[align=c, width=0.16\linewidth]{imgs/flower/2_unseen_86.jpg}
%         & \includegraphics[align=c, width=0.16\linewidth]{imgs/flower/4_unseen_86.jpg}
%         & \includegraphics[align=c, width=0.16\linewidth]{imgs/flower/5_unseen_86.jpg} \\
%     \end{tabular}
%     \caption{\textbf{Unseen flower category generation.} Qualitative comparison between real images, SN-GAN, and KG-GAN. These two flower categories (Orange Dahlia and Stemless Gentian) are seen to SN-GAN while unseen to KG-GAN.}
%     \label{fig:flower0}
% \end{center}
% \end{figure}

\begin{figure}[htbp]
\tabcolsep=1pt
\renewcommand\arraystretch{5.1}
\begin{center}
    \begin{tabular}{c c c c c}
        \scriptsize{Real images} & \scriptsize{SN-GAN} & \scriptsize{One-hot KG-GAN} & \scriptsize{KG-GAN w/o $\LL_{se}$} & \scriptsize{KG-GAN} \\  [-16pt]
        \includegraphics[align=c, width=0.2\linewidth]{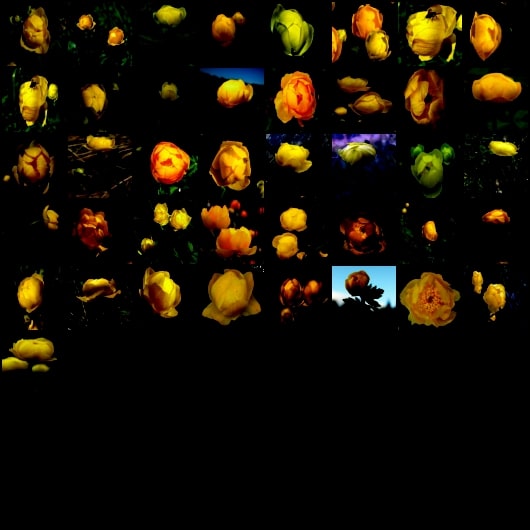}
        & \includegraphics[align=c, width=0.2\linewidth]{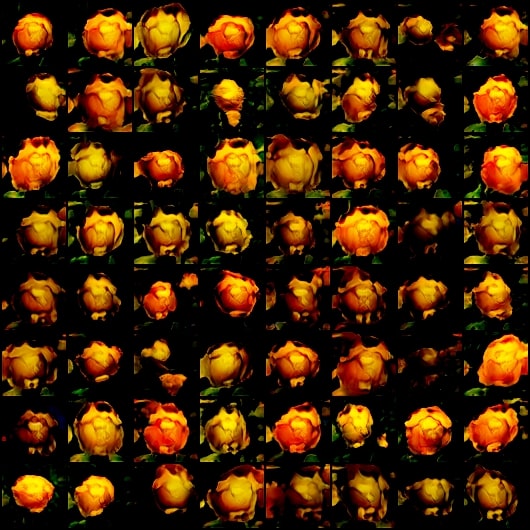}
        & \includegraphics[align=c, width=0.2\linewidth]{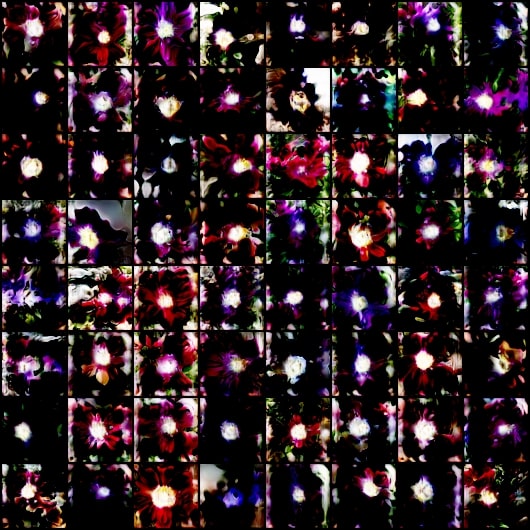}
        & \includegraphics[align=c, width=0.2\linewidth]{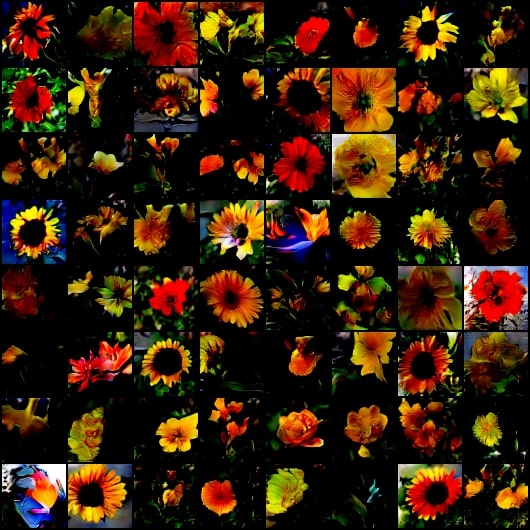}
        & \includegraphics[align=c, width=0.2\linewidth]{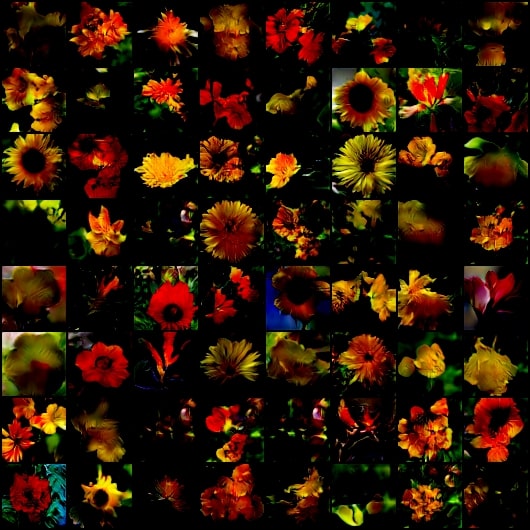} \\

        \includegraphics[align=c, width=0.2\linewidth]{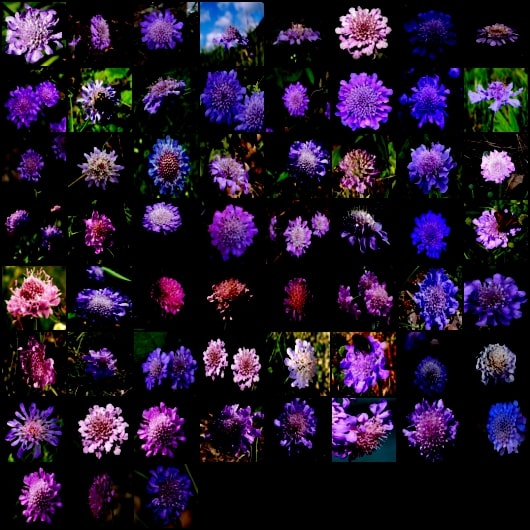}
        & \includegraphics[align=c, width=0.2\linewidth]{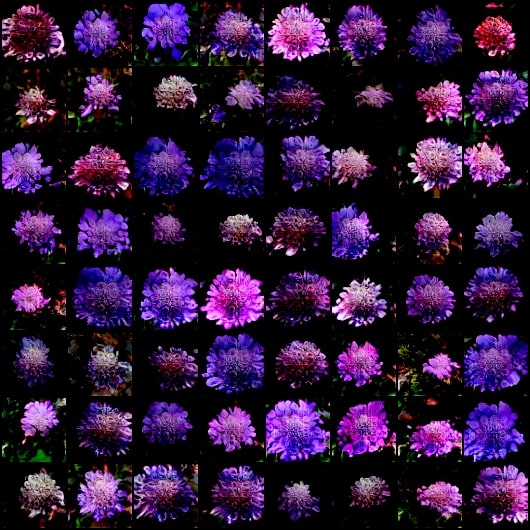}
        & \includegraphics[align=c, width=0.2\linewidth]{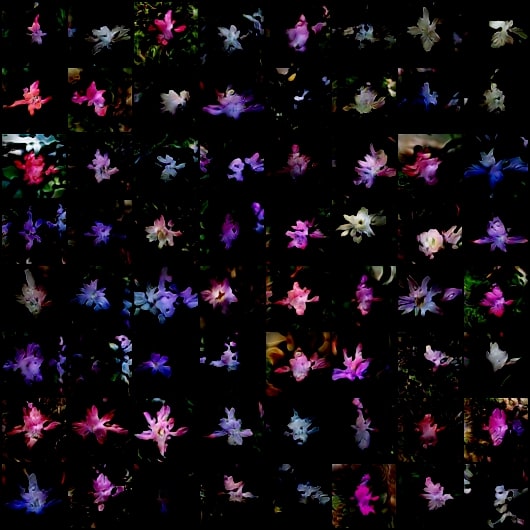}
        & \includegraphics[align=c, width=0.2\linewidth]{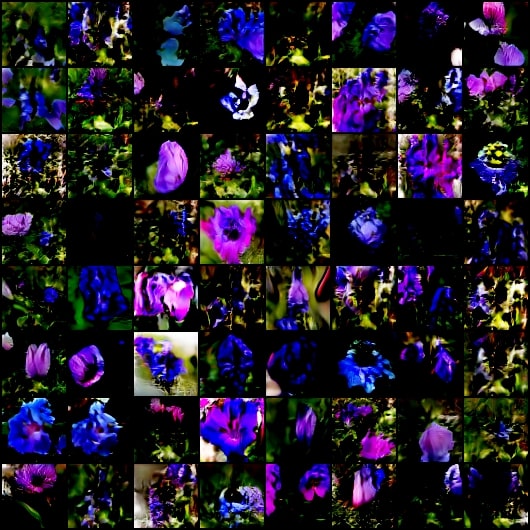}
        & \includegraphics[align=c, width=0.2\linewidth]{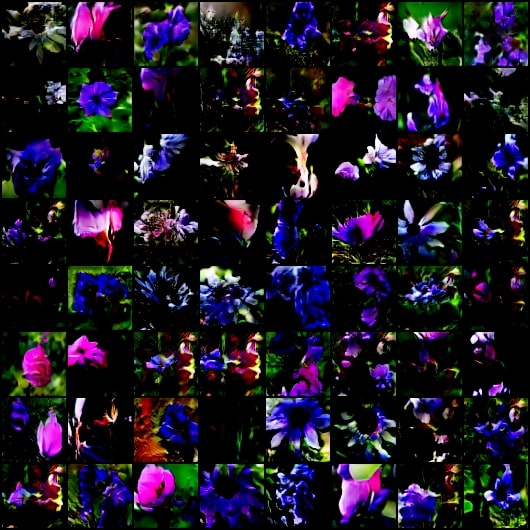} \\

        \includegraphics[align=c, width=0.2\linewidth]{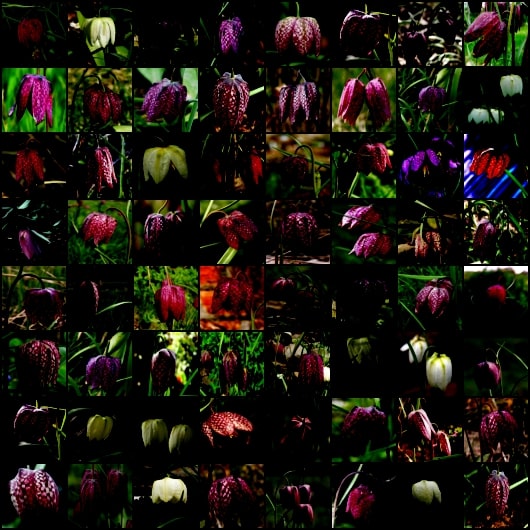}
        & \includegraphics[align=c, width=0.2\linewidth]{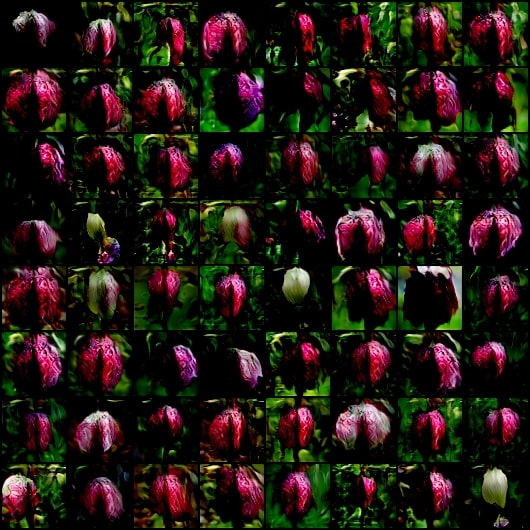}
        & \includegraphics[align=c, width=0.2\linewidth]{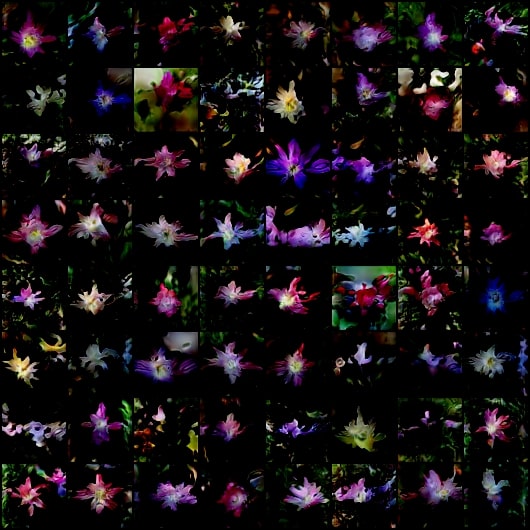}
        & \includegraphics[align=c, width=0.2\linewidth]{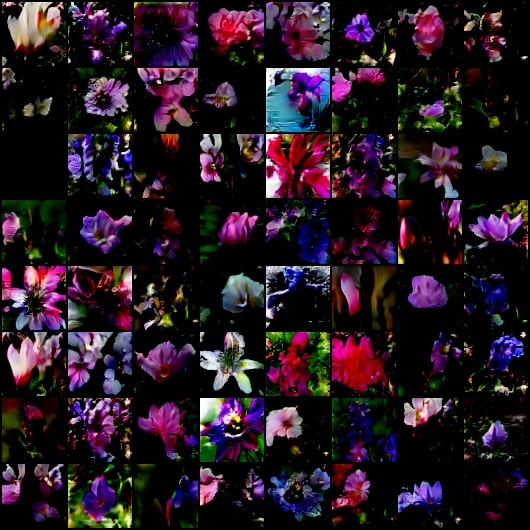}
        & \includegraphics[align=c, width=0.2\linewidth]{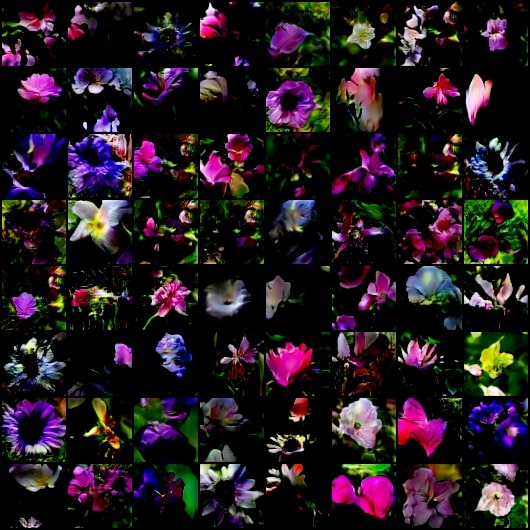} \\

        \includegraphics[align=c, width=0.2\linewidth]{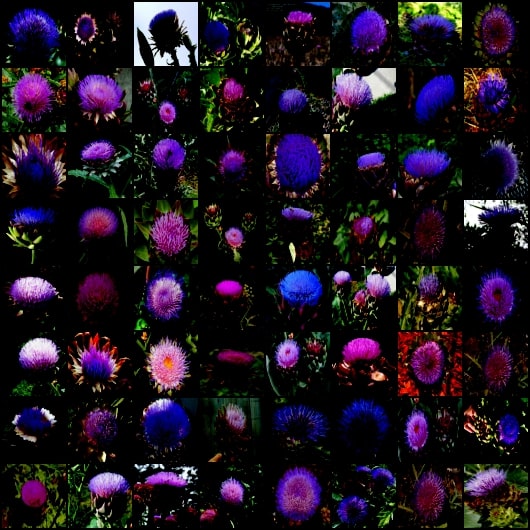}
        & \includegraphics[align=c, width=0.2\linewidth]{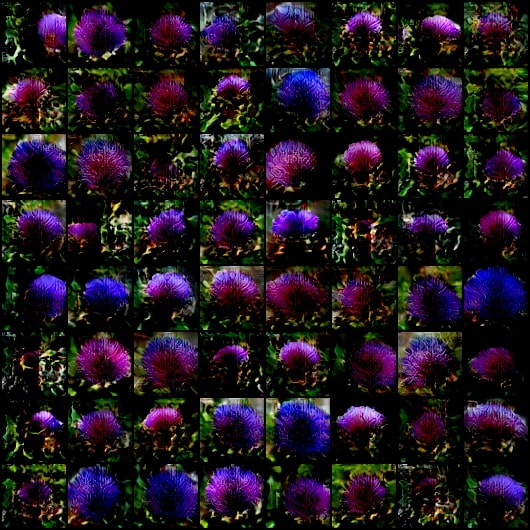}
        & \includegraphics[align=c, width=0.2\linewidth]{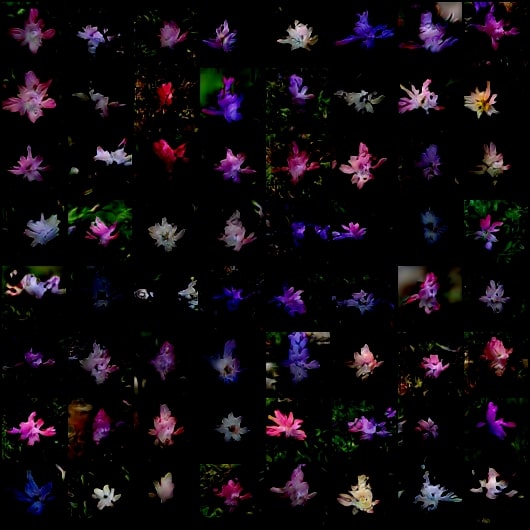}
        & \includegraphics[align=c, width=0.2\linewidth]{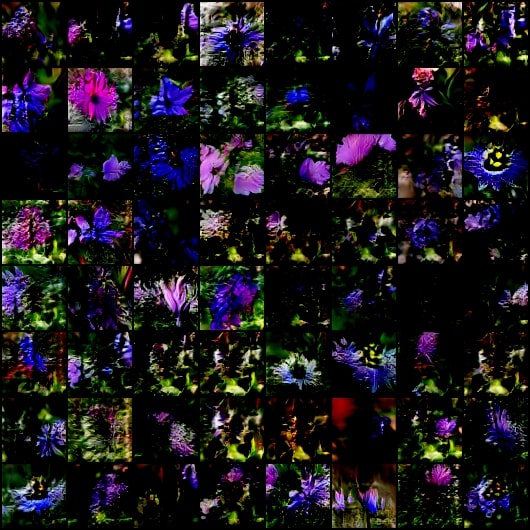}
        & \includegraphics[align=c, width=0.2\linewidth]{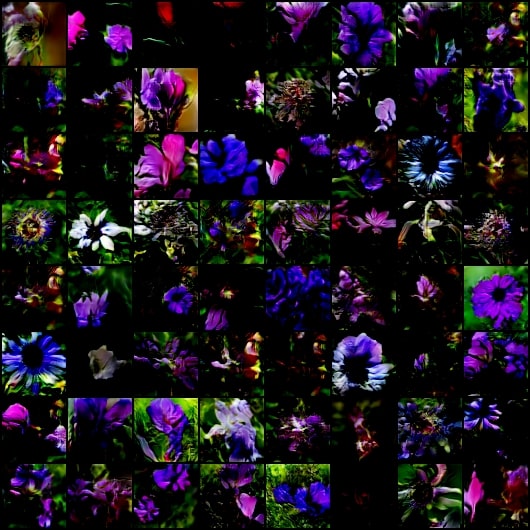} \\

        \includegraphics[align=c, width=0.2\linewidth]{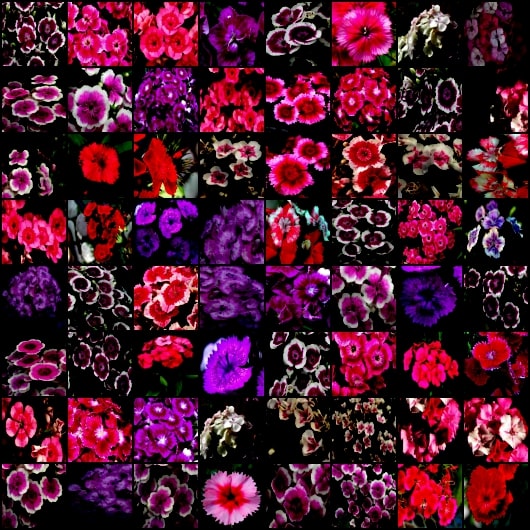}
        & \includegraphics[align=c, width=0.2\linewidth]{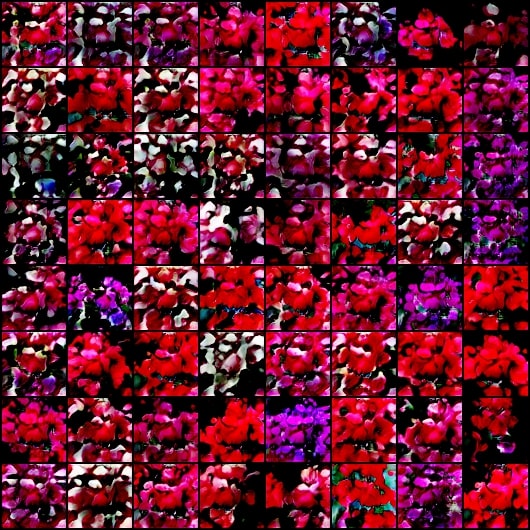}
        & \includegraphics[align=c, width=0.2\linewidth]{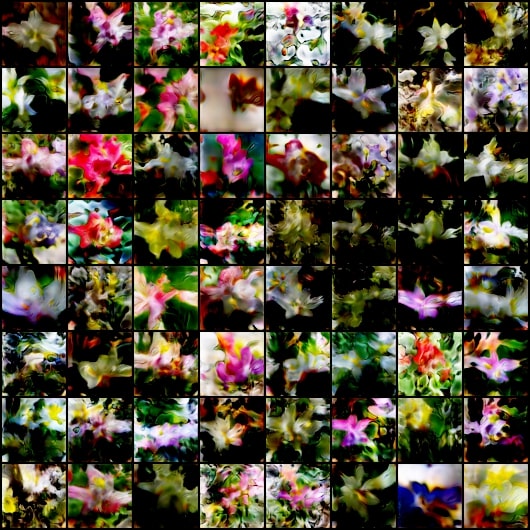}
        & \includegraphics[align=c, width=0.2\linewidth]{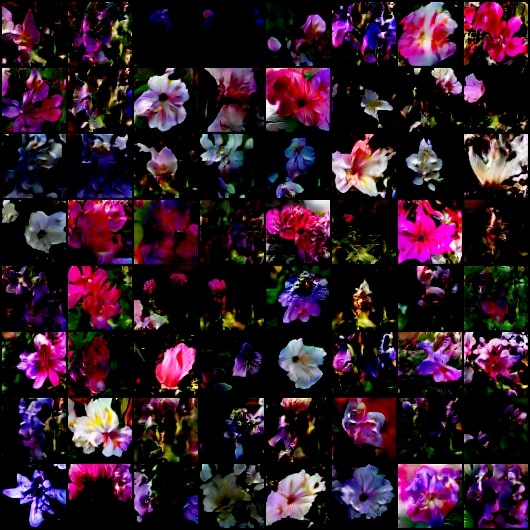}
        & \includegraphics[align=c, width=0.2\linewidth]{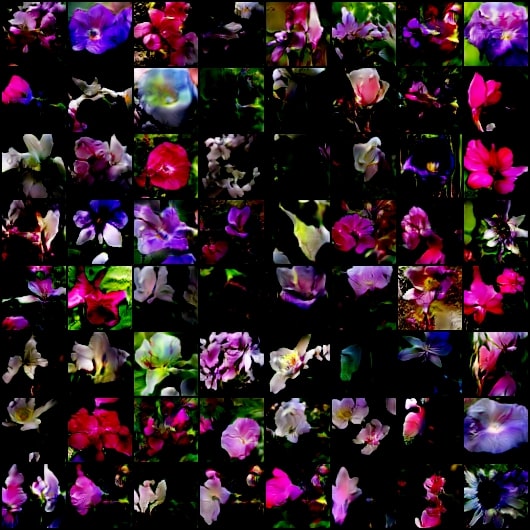} \\

    \end{tabular}
    \caption{Bougainvillea, Canterbury Bells, CapeFlower, Common Dandelion, and Corn Poppy. Please zoom in to see the details.}
    \label{fig:flower1}
\end{center}
\end{figure}

\begin{figure}[htbp]
\tabcolsep=1pt
\renewcommand\arraystretch{5.2}
\begin{center}
    \begin{tabular}{c c c c c}
        \scriptsize{Real images} & \scriptsize{SN-GAN} & \scriptsize{One-hot KG-GAN} & \scriptsize{KG-GAN w/o $\LL_{se}$} & \scriptsize{KG-GAN} \\  [-16pt]
        \includegraphics[align=c, width=0.2\linewidth]{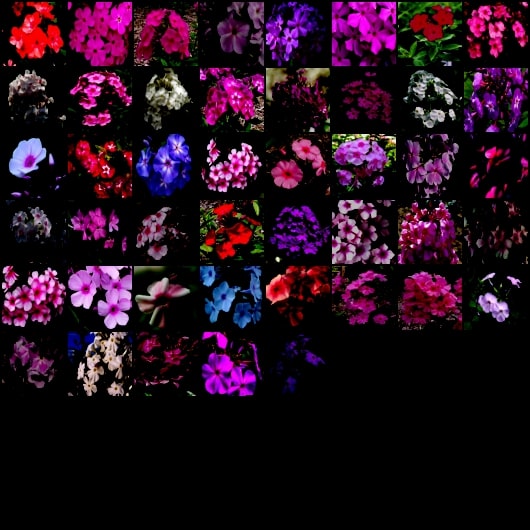}
        & \includegraphics[align=c, width=0.2\linewidth]{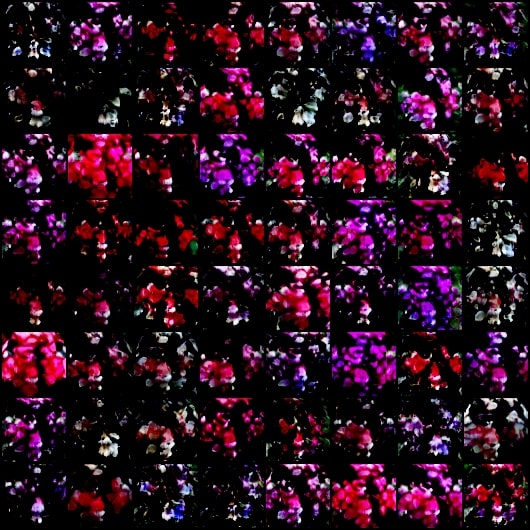}
        & \includegraphics[align=c, width=0.2\linewidth]{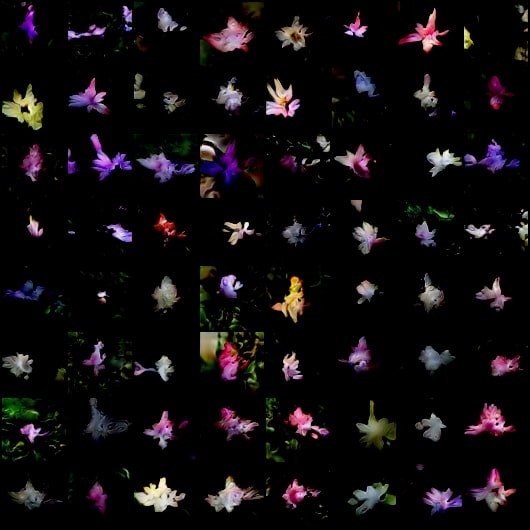}
        & \includegraphics[align=c, width=0.2\linewidth]{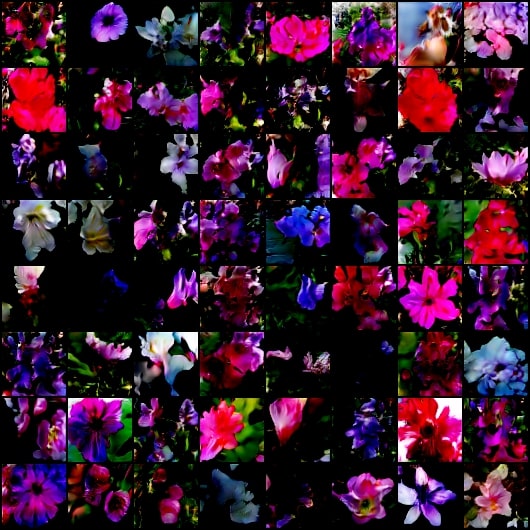}
        & \includegraphics[align=c, width=0.2\linewidth]{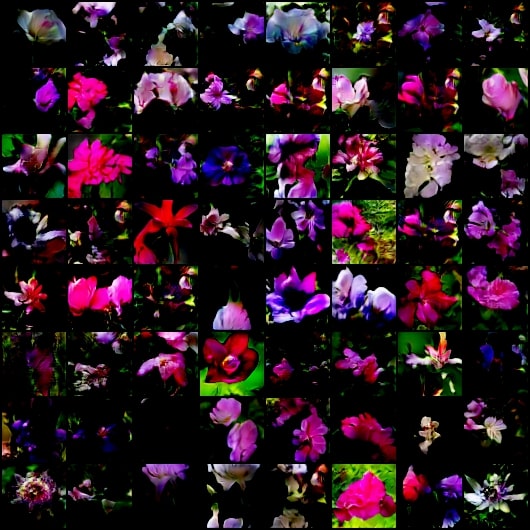} \\

        \includegraphics[align=c, width=0.2\linewidth]{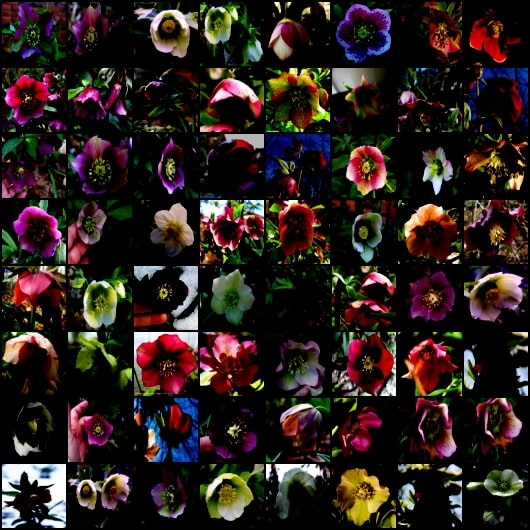}
        & \includegraphics[align=c, width=0.2\linewidth]{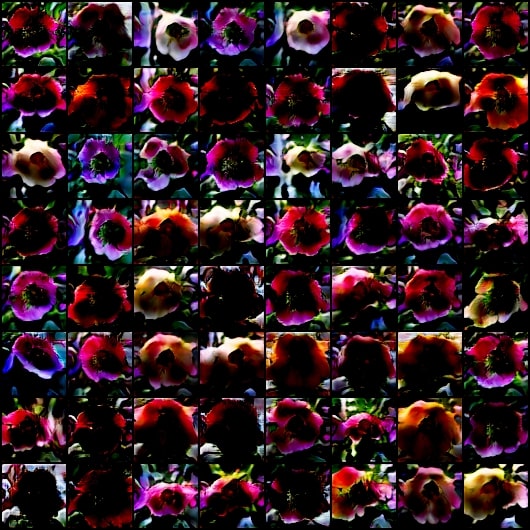}
        & \includegraphics[align=c, width=0.2\linewidth]{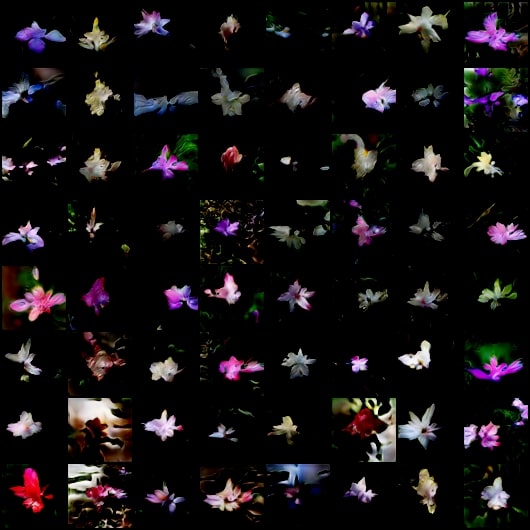}
        & \includegraphics[align=c, width=0.2\linewidth]{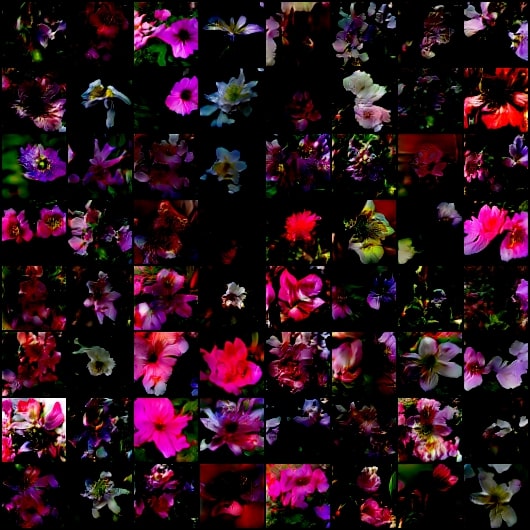}
        & \includegraphics[align=c, width=0.2\linewidth]{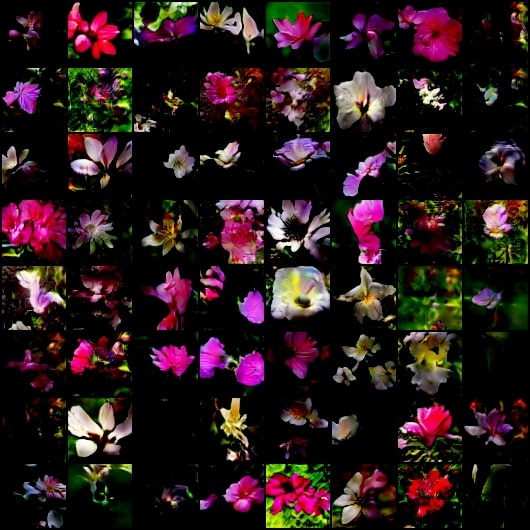} \\

        \includegraphics[align=c, width=0.2\linewidth]{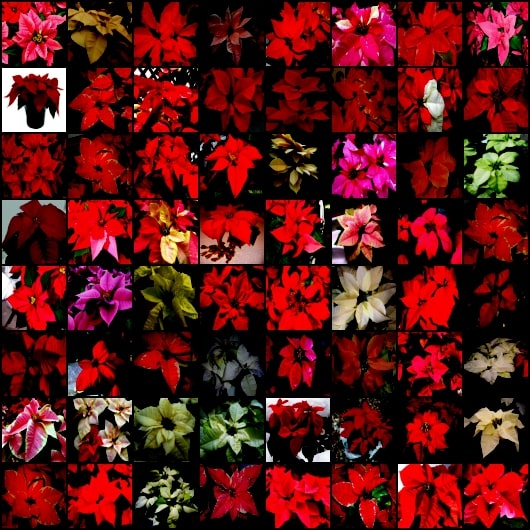}
        & \includegraphics[align=c, width=0.2\linewidth]{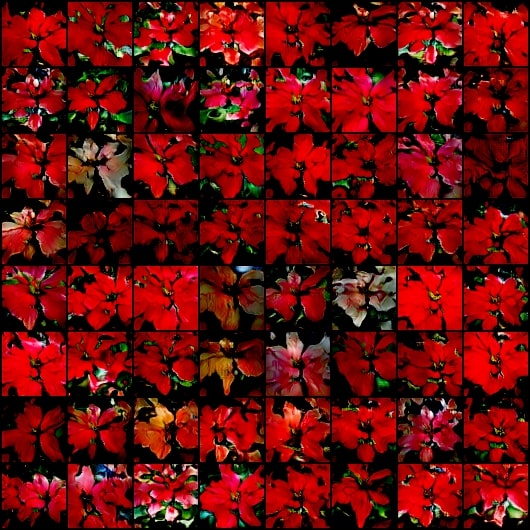}
        & \includegraphics[align=c, width=0.2\linewidth]{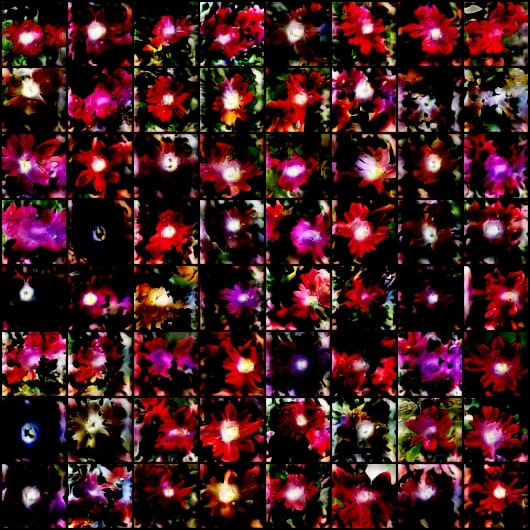}
        & \includegraphics[align=c, width=0.2\linewidth]{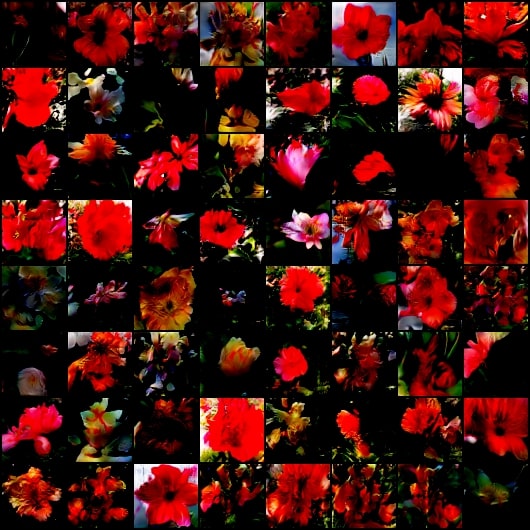}
        & \includegraphics[align=c, width=0.2\linewidth]{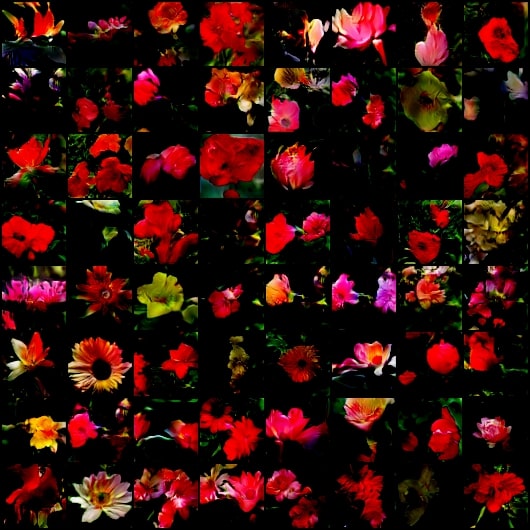} \\

        \includegraphics[align=c, width=0.2\linewidth]{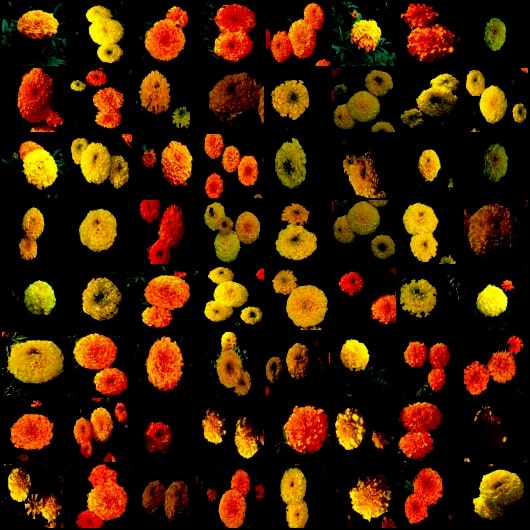}
        & \includegraphics[align=c, width=0.2\linewidth]{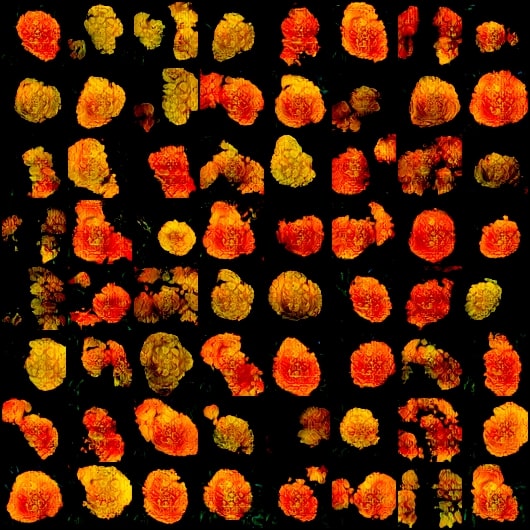}
        & \includegraphics[align=c, width=0.2\linewidth]{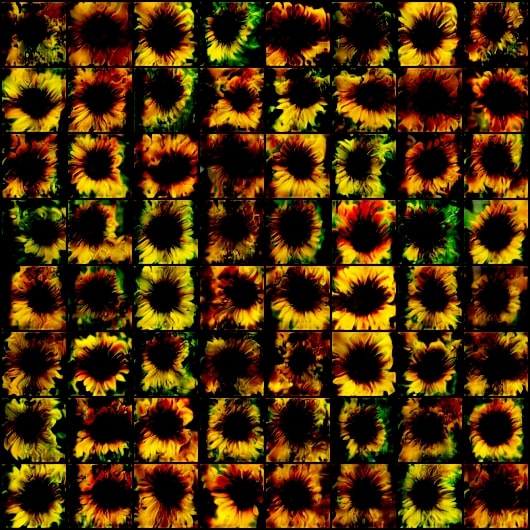}
        & \includegraphics[align=c, width=0.2\linewidth]{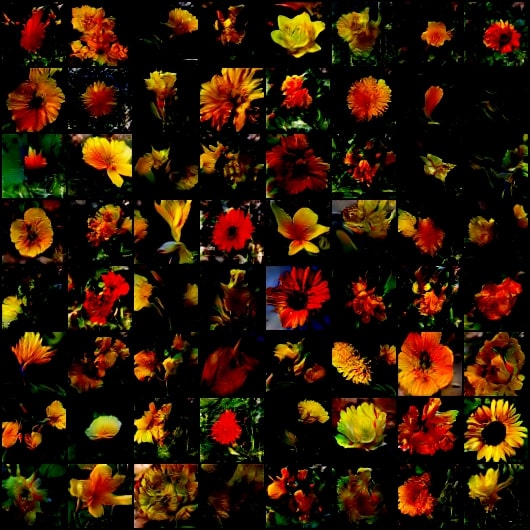}
        & \includegraphics[align=c, width=0.2\linewidth]{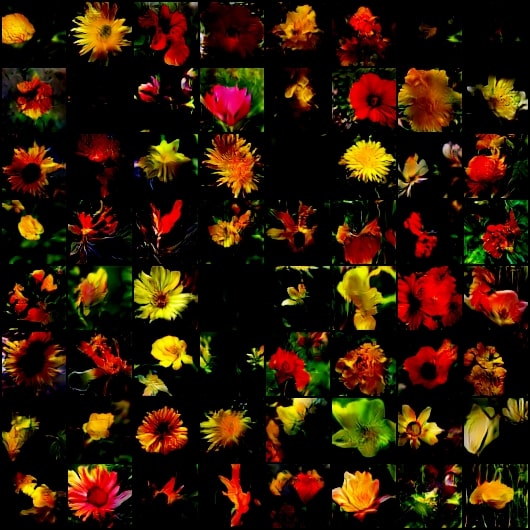} \\

        \includegraphics[align=c, width=0.2\linewidth]{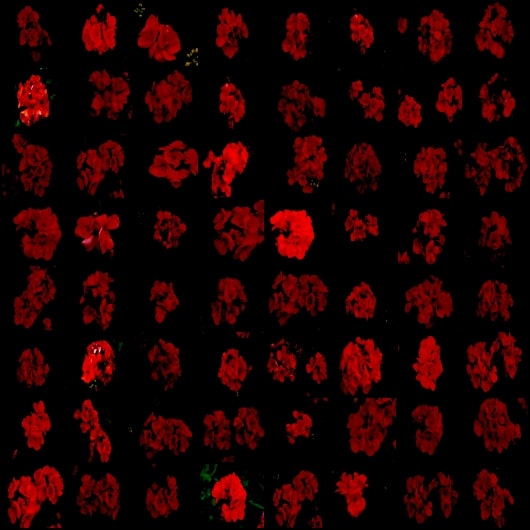}
        & \includegraphics[align=c, width=0.2\linewidth]{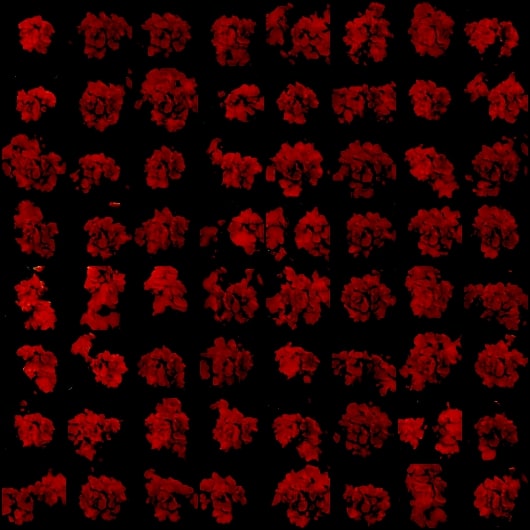}
        & \includegraphics[align=c, width=0.2\linewidth]{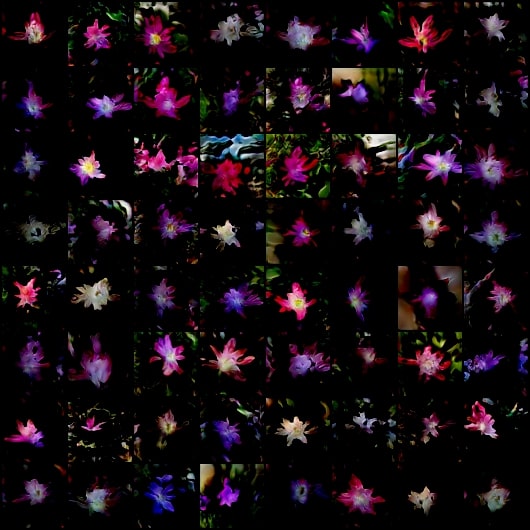}
        & \includegraphics[align=c, width=0.2\linewidth]{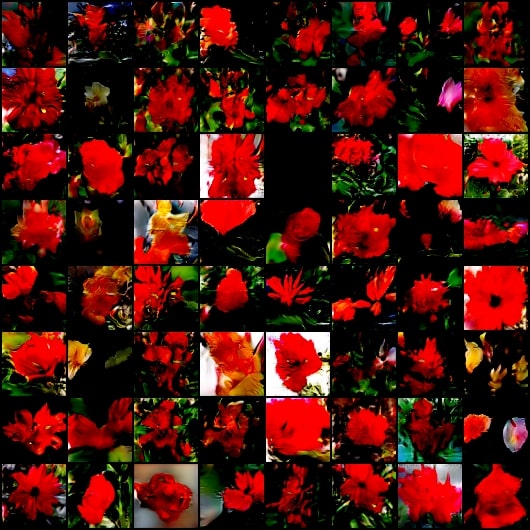}
        & \includegraphics[align=c, width=0.2\linewidth]{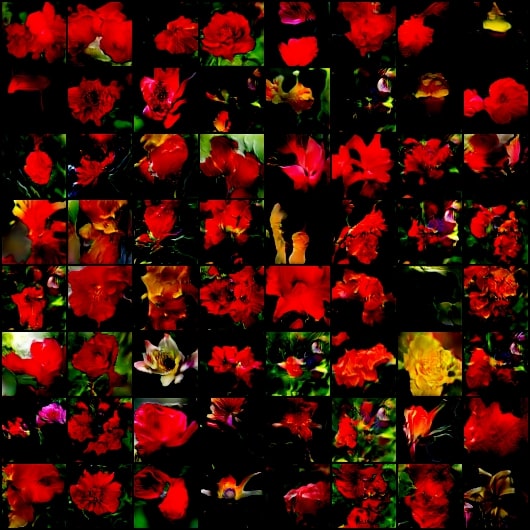} \\

    \end{tabular}
    \caption{Daffodil, Gaura, Globe Thistle, Great Masterwort, and Marigold. Please zoom in to see the details.}
    \label{fig:flower2}
\end{center}
\end{figure}

\begin{figure}[htbp]
\tabcolsep=1pt
\renewcommand\arraystretch{5.2}
\begin{center}
    \begin{tabular}{c c c c c}
        \scriptsize{Real images} & \scriptsize{SN-GAN} & \scriptsize{One-hot KG-GAN} & \scriptsize{KG-GAN w/o $\LL_{se}$} & \scriptsize{KG-GAN} \\  [-16pt]

        \includegraphics[align=c, width=0.2\linewidth]{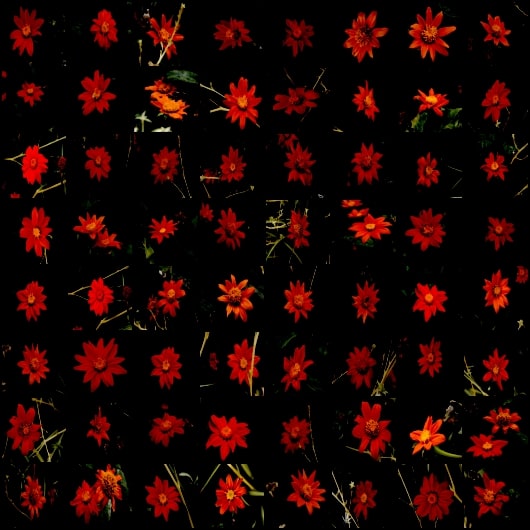}
        & \includegraphics[align=c, width=0.2\linewidth]{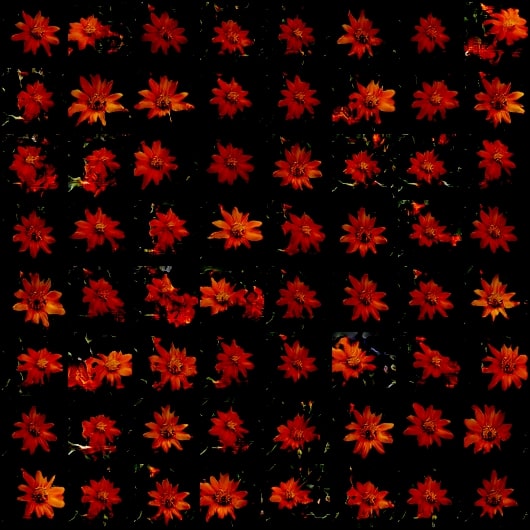}
        & \includegraphics[align=c, width=0.2\linewidth]{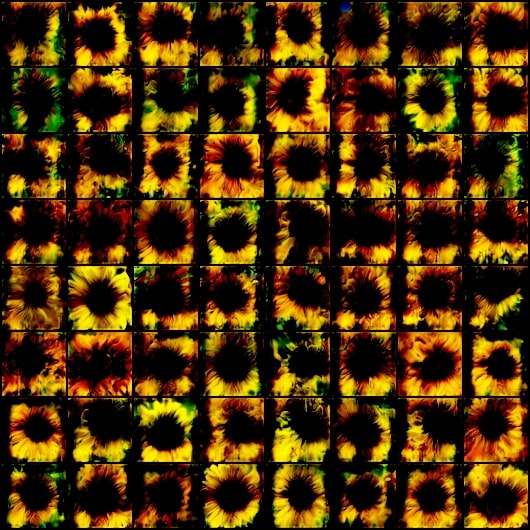}
        & \includegraphics[align=c, width=0.2\linewidth]{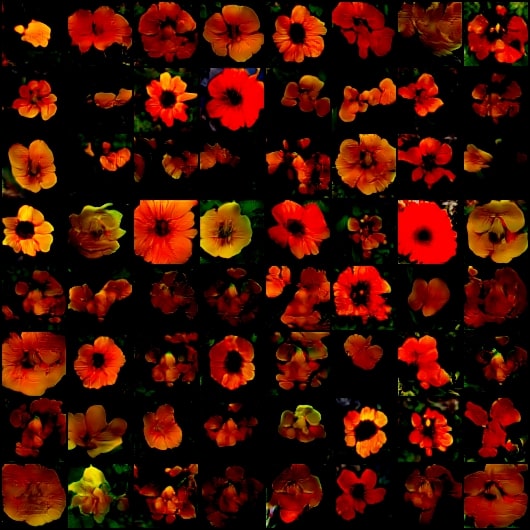}
        & \includegraphics[align=c, width=0.2\linewidth]{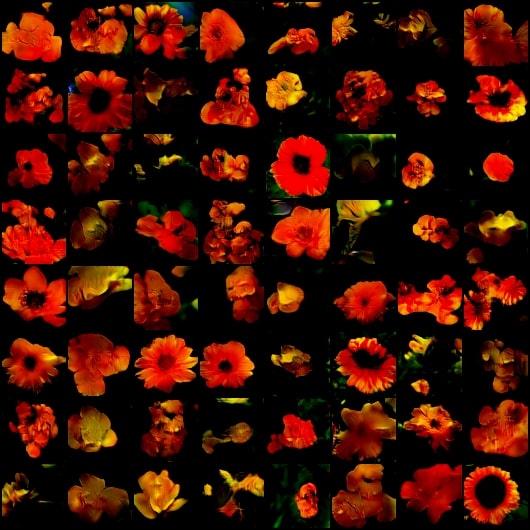} \\

        \includegraphics[align=c, width=0.2\linewidth]{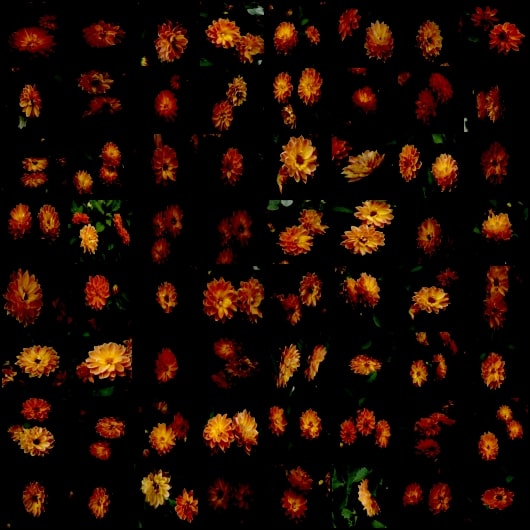}
        & \includegraphics[align=c, width=0.2\linewidth]{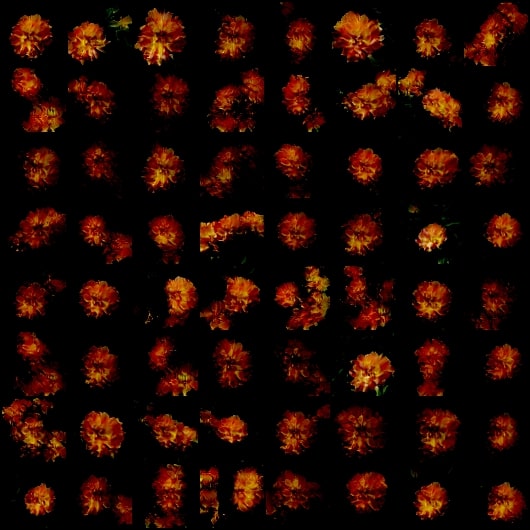}
        & \includegraphics[align=c, width=0.2\linewidth]{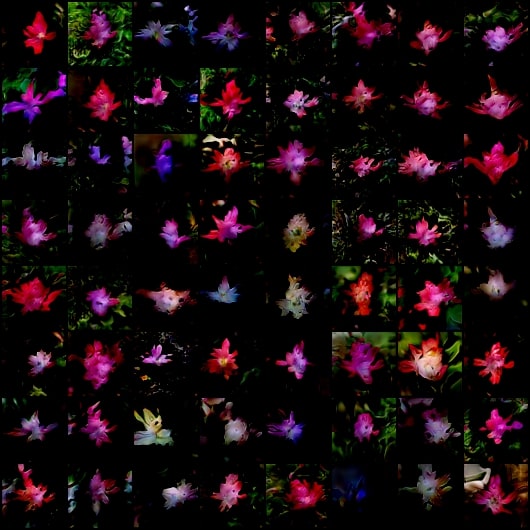}
        & \includegraphics[align=c, width=0.2\linewidth]{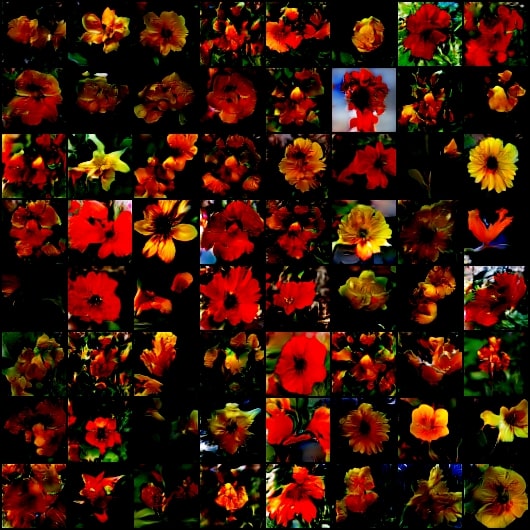}
        & \includegraphics[align=c, width=0.2\linewidth]{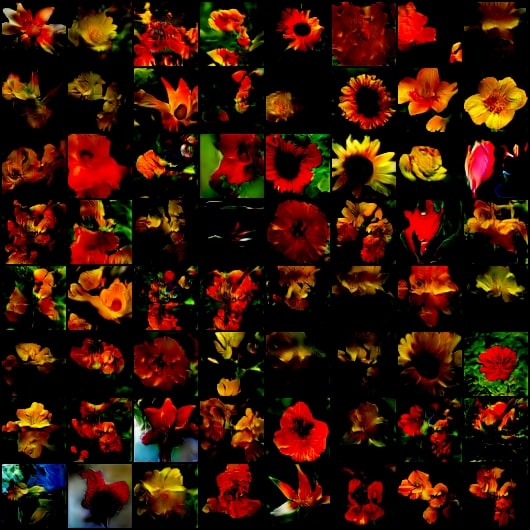} \\

        \includegraphics[align=c, width=0.2\linewidth]{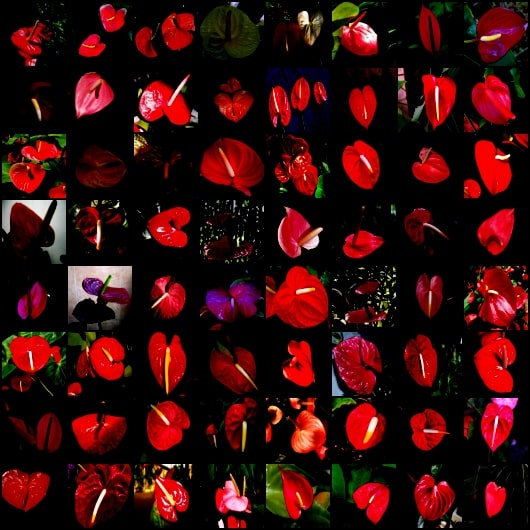}
        & \includegraphics[align=c, width=0.2\linewidth]{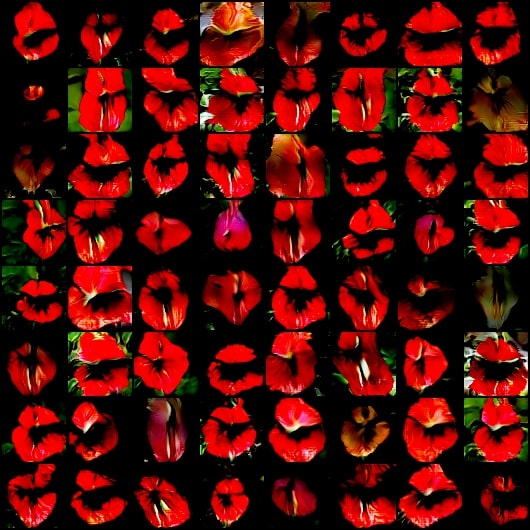}
        & \includegraphics[align=c, width=0.2\linewidth]{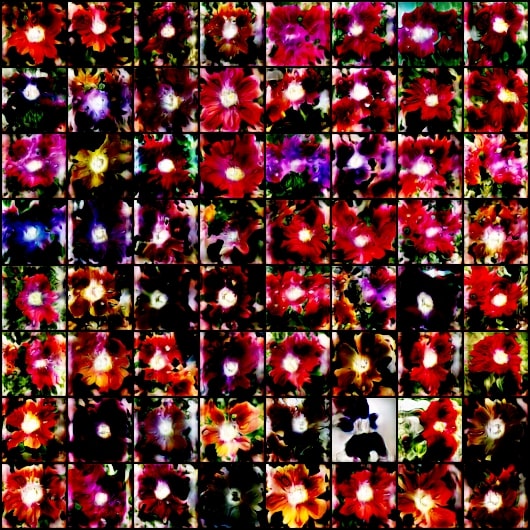}
        & \includegraphics[align=c, width=0.2\linewidth]{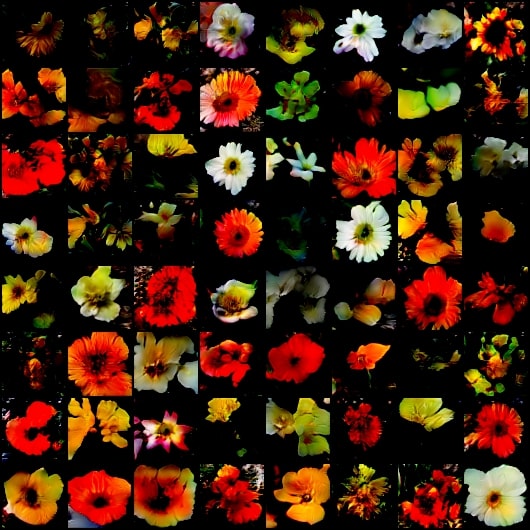}
        & \includegraphics[align=c, width=0.2\linewidth]{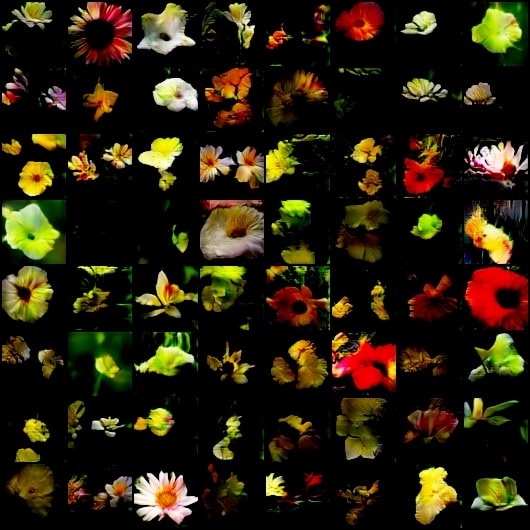} \\

        \includegraphics[align=c, width=0.2\linewidth]{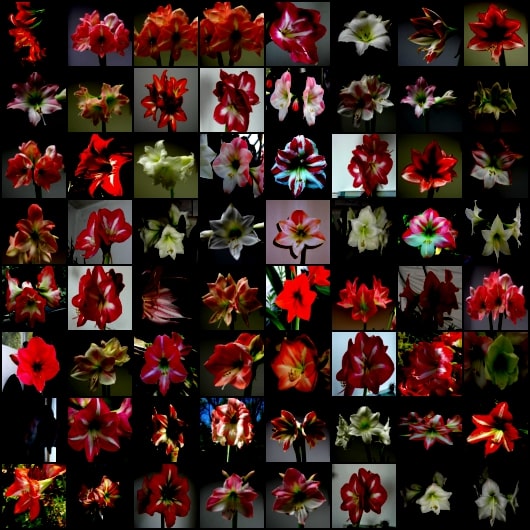}
        & \includegraphics[align=c, width=0.2\linewidth]{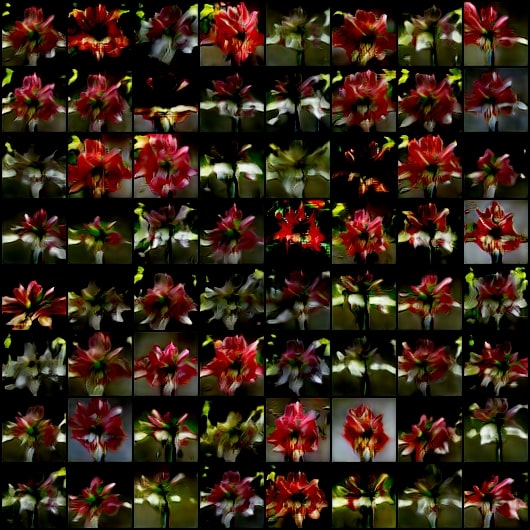}
        & \includegraphics[align=c, width=0.2\linewidth]{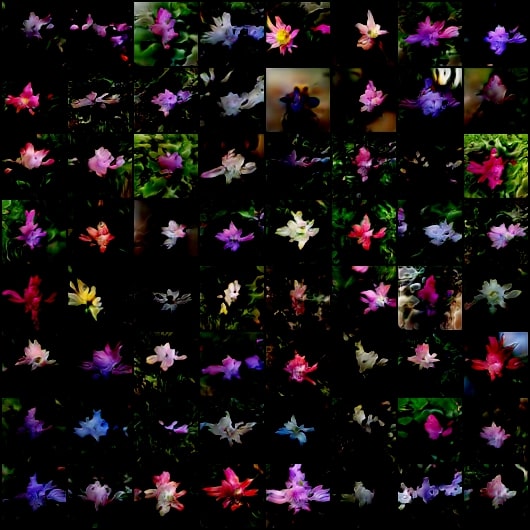}
        & \includegraphics[align=c, width=0.2\linewidth]{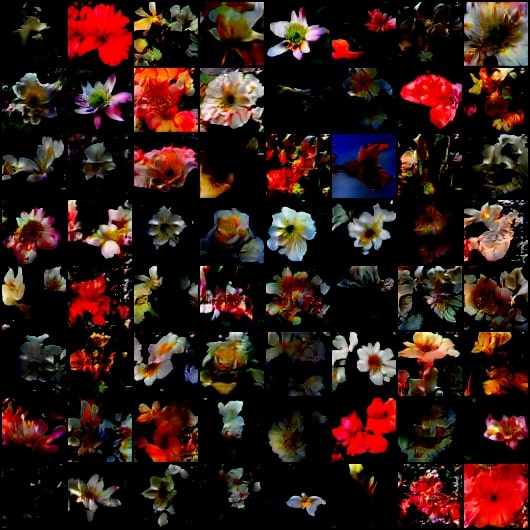}
        & \includegraphics[align=c, width=0.2\linewidth]{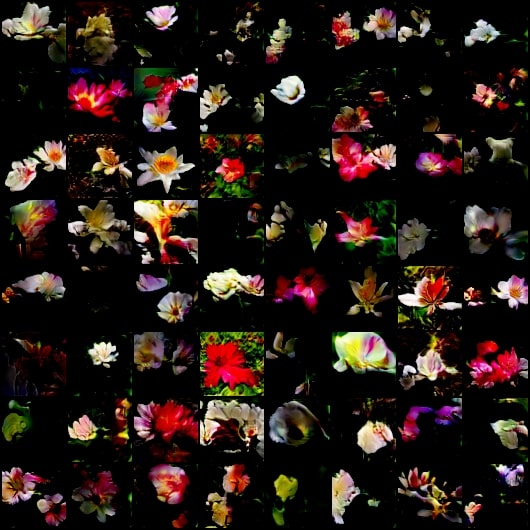} \\

        \includegraphics[align=c, width=0.2\linewidth]{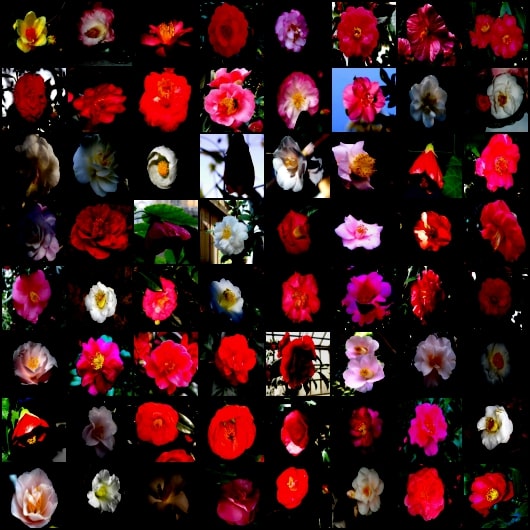}
        & \includegraphics[align=c, width=0.2\linewidth]{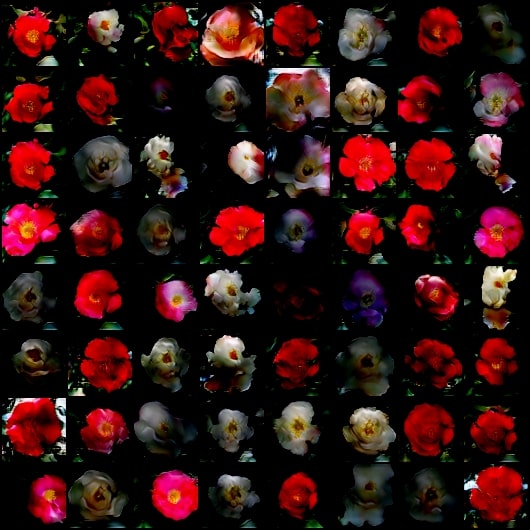}
        & \includegraphics[align=c, width=0.2\linewidth]{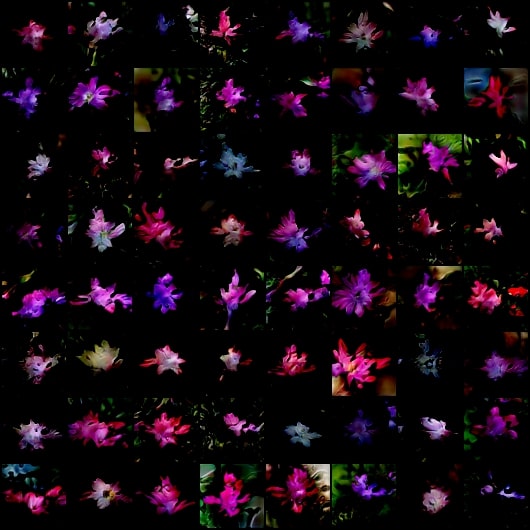}
        & \includegraphics[align=c, width=0.2\linewidth]{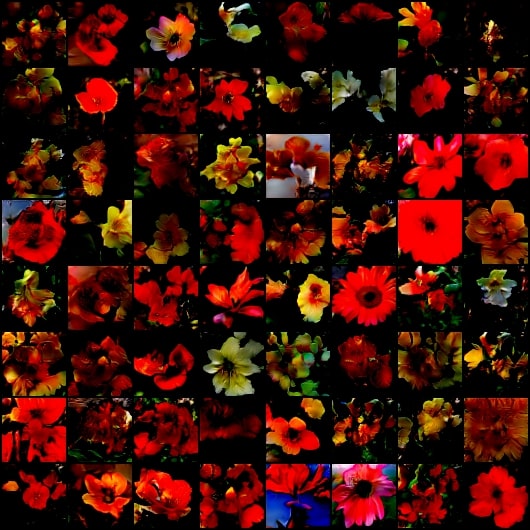}
        & \includegraphics[align=c, width=0.2\linewidth]{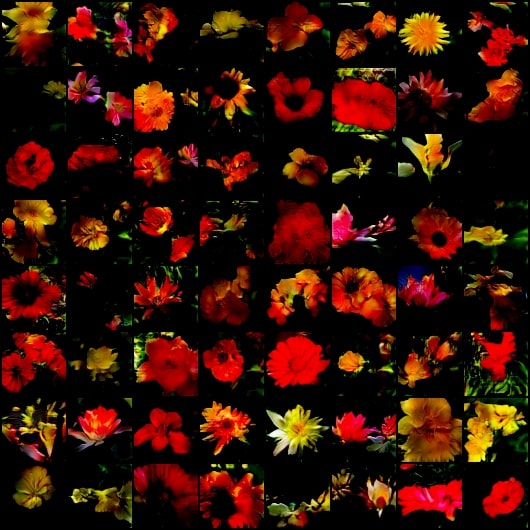} \\

    \end{tabular}
    \caption{Mexican Aster, Mexican Petunia, Ruby-lipped Cattleya, Thorn Apple, and TrumpetCreeper. Please zoom in to see the details.}
    \label{fig:flower3}
\end{center}
\end{figure}